\documentclass{article}

\usepackage{microtype}
\usepackage{graphicx}
\usepackage{subcaption}
\usepackage{booktabs} % for professional tables

\usepackage{hyperref}

\usepackage[accepted]{icml2026}

\usepackage{amsmath}
\usepackage{amssymb}
\usepackage{mathtools}
\usepackage{amsthm}
\usepackage{multirow}

\usepackage[capitalize,noabbrev]{cleveref}

\theoremstyle{plain}
\newtheorem{theorem}{Theorem}[section]

\theoremstyle{definition}
\newtheorem{definition}[theorem]{Definition}

\theoremstyle{remark}

\usepackage{etoolbox}

\usepackage[textsize=tiny]{todonotes}

\icmltitlerunning{Cross-Modal Knowledge Distillation without Paired Data}

\begin{document}

\twocolumn[

  \icmltitle{Cross-Modal Knowledge Distillation without Paired Data:\\ Theoretical Foundation and Algorithm}
  %\icmltitle{Principal Approach for Unpaired Cross-Modality Knowledge Distillation via Optimizing the Interaction between Feature Alignment and Label Alignment}



  \icmlsetsymbol{equal}{*}

  \begin{icmlauthorlist}
    \icmlauthor{Trong Khiem Tran}{equal,wsu,hust}
    \icmlauthor{Anh Duc Chu}{equal,hust}
    \icmlauthor{Quang Hung Pham}{hust}
    \icmlauthor{Phi Le Nguyen}{hust}
    \icmlauthor{Trong Nghia Hoang}{wsu}
  \end{icmlauthorlist}

  \icmlaffiliation{hust}{School of Information and Communications Technology, Hanoi University of Science and Technology, Hanoi, Vietnam}
  \icmlaffiliation{wsu}{School of Electrical Engineering and Computer Science, Washington State University, Pullman, US}

  \icmlcorrespondingauthor{Trong Nghia Hoang}{trongnghia.hoang@wsu.edu}
  \icmlcorrespondingauthor{Trong Khiem Tran}{khiem.tran@wsu.edu}
  \icmlkeywords{Machine Learning, ICML}

  \vskip 0.3in
]



\printAffiliationsAndNotice{}  % no special notice (required even if empty)

\begin{abstract}
Cross-modal knowledge distillation (CMKD) studies how a (large) teacher model trained on one type of data (e.g., images) can guide a (smaller) student model building on another type of data (e.g., text/audio).~Existing CMKD methods often require paired multi-modal data with aligned semantics, but obtaining such paired data are often costly and impractical.~To mitigate this limitation, we develop a new CMKD framework for the more challenging setting where paired data are unavailable. In particular, we establish a cross-modal distributional relationship between teacher and student models which reveals two fundamental quantities governing effective distillation: feature alignment and label alignment.~These quantities characterize semantic discrepancy between modalities at the levels of representation and prediction distributions, respectively.~Motivated by this insight, we propose a principled framework, with theoretical guarantees, that enables effective cross-modal knowledge distillation by aligning distributions rather than individual samples.~Extensive experiments across a wide range of multimodal benchmarks show that our framework is highly effective in both unpaired and paired data settings, improving significantly over prior work.

\end{abstract}

\section{Introduction}
\label{sec:intro}
Knowledge distillation (KD) was first introduced by \citep{Buci2006model} and \citep{hinton2015distillingknowledgeneuralnetwork} as a mechanism for transferring predictive knowledge from a complex teacher model to a simpler student model.~The key motivation is that highly over-parameterized models are often beneficial during training, as the additional capacity provides flexibility for making mistakes and correcting them.~The learned knowledge however can often be compressed into a significantly smaller model with minimal loss in predictive performance, making deployment much more efficient.~This can be achieved by making the student model mimic the behavior of the teacher model via matching soft predictions.

In classical in-modal settings, both the teacher and student models operate on the same input representation and data modalities.~An existing rich literature has advanced in-modal knowledge distillation by improving how knowledge is transferred from teacher to student models.~These include contrastive losses for representation alignment~\citep{Tian2020Contrastive}, variance reduction via Bayes-optimal teachers~\citep{menon21stastkd}, gradient-aware adaptive distillation~\citep{Zhu2021SCKD}, decoupled treatment of target and non-target classes~ \citep{Zhao2022DKD}, multi-level feature review mechanisms~\citep{Chen2021Review}, and relation-based losses capturing inter- and intra-class structure~\citep{huang2022knowledgedistillationstrongerteacher, Yang2023MLKD}.~A more comprehensive review of classical in-modal KD methods is provided in Section~\ref{sec:related_work_uni_kd}.

In contrast, cross-modal knowledge distillation (CMKD) considers more challenging and practical settings where teacher and student models operate on different data modalities, such as transferring knowledge from image-based models to text- or audio-based models.~In its distillation process, the teacher and student inputs correspond to different modalities of the same underlying instance, which typically requires paired multimodal data during training. Recent advances have demonstrated the broad applicability of CMKD across diverse domains, including selective bidirectional distillation for bridging modality gaps (C2KD)~\citep{Huo2024C2KD}, self-supervised representation learning from unlabeled videos~\citep{sarkar2023xkdcrossmodalknowledgedistillation}, vision–language self-distillation via cross-attention mechanisms (COSMOS)~\citep{kim2025cosmoscrossmodalityselfdistillationvision}, and weakly paired transfer between microscopy images and transcriptomics data (XKD)~\citep{bendidi2025a}.~Despite this progress, existing CMKD methods remain largely dependent on paired or weakly paired multimodal data with sample-level correspondence, which is often costly, difficult to obtain, or impractical in real-world settings where modalities are collected independently or asynchronously.~This raises the following fundamental challenge: 

%In its distillation process, the input of the teacher and the student comes from different modalities of a single instance, thus requiring the paired data for training datasets. Recent advances demonstrate the broad applicability of cross-modal knowledge distillation across practical domains, including selective bidirectional distillation to bridge modality gaps (C2KD; \citealp{Huo2024C2KD}), self-supervised representation learning from unlabeled videos \citep{sarkar2023xkdcrossmodalknowledgedistillation}, vision–language self-distillation via text cropping and cross-attention (COSMOS; \citealp{kim2025cosmoscrossmodalityselfdistillationvision}), and weakly paired cross-modal transfer from microscopy images to transcriptomics (XKD; \citealp{bendidi2025a}). Despite its broad applicability, cross-modal knowledge distillation (CMKD) is still largely constrained by the requirement of paired or weakly paired multimodal data, which significantly limits its scalability and deployment in real-world scenarios where modalities are often collected independently or asynchronously. This raises a fundamental challenge: 

\noindent {\bf How can we distill knowledge from the teacher model to the student without using sample-level pairing ?}%\vspace{1mm}

A natural extension to the unpaired setting is to adapt feature-based knowledge distillation methods by replacing sample-level feature matching with distribution-level feature alignment.~However, prior work \citep{Huo2024C2KD} has shown that direct adoption of conventional feature-based distillation methods, such as FitNet \citep{romero2015fitnetshintsdeepnets} and ReviewKD \citep{Chen2021Review}, to cross-modal settings remains ineffective even when paired data are available.~This limitation arises because effective cross-modal transfer requires aligning not only representation structures but also predictive semantics across modalities.~In the unpaired setting, where semantic sample-level correspondence is entirely absent, this challenge becomes even more severe, as further confirmed by our experimental results (Section~\ref{sec:unpaired_results}).

%The naive solution for this challenge is Feature-based Knowledge Distillation, which can be modified to align the representation of the teacher model and the student model on the shared latent space at the data batch-level instead of the data-point level. In this way, the modified Feature-based Knowledge Distillation can run on the unpaired data setting for distillation. However, \citet{Huo2024C2KD} shows that, even in the paired data scenario, directly using feature-baseddistillation methods, such  as FitNet~\citep{romero2015fitnetshintsdeepnets} and Review~\citep{Chen2021Review}, for CMKD is unreasonable because of significant divergence between teacher and student. Extend to the unpaired data scenario, directly using feature-based distillation for CMKD is even more unreasonable due to the additional challenge of the lack of semantic data-pairing. Which leaves the research gap on using Knowledge Distillation in the unpaired data scenario. 

To address this challenge, we introduce a principled framework that establishes a provable bound on the generalized student error via two fundamental quantities:~(i) \textbf{feature alignment} and (ii) \textbf{label alignment}.~These quantities characterize semantic discrepancy across modalities at the levels of representation and prediction distributions via  measuring the cross-modal gap between the teacher model and the student, given a shared representation space.~Larger cross-modal discrepancies make effective knowledge transfer substantially more difficult and can lead to suboptimal distillation.~This characterization provides actionable guidance for minimizing cross-modal discrepancy, leading to a theoretically grounded framework for algorithm design. Our key contributions are summarized as follows:

%To bridge this gap, we introduce a principled framework that establishes a provable bound on the generalized student error,  capturing the semantic discrepancy between modalities at the level of representation distributions and prediction distributions via \textbf{feature alignment} and \textbf{label alignment} concepts, respectively. Intuitively, feature alignment and label alignment measure the cross-modal gap between the teacher model and the student, given a shared representation space. A large cross-modality gap means the student model is more challenging to mimic the teacher's behavior, thus easily falls into the suboptimal regime. This decomposition provides actionable insights into how the gap between the teacher model and the student model should be optimized, offering a theoretically grounded guide for algorithm design.~The above is substantiated with the following contributions:

%\textbf{1.~Theoretical Analysis}.~We develop a theoretical bound that decomposes the generalized student error into: (i) the teacher error, which serves as a fixed overhead;  (ii) the representation distributional distance between the teacher and the student (\textbf{feature alignment}); (iii) the prediction distributional distance between the teacher and the student (\textbf{label alignment}). To the best of our knowledge, this is the first generalization bound that captures the influence of both the teacher model’s quality and the interaction between feature alignment and label alignment on the student performance in setting CMKD (Section~\ref{sec:theory}).

\textbf{1.~Theoretical Analysis}.~We develop a theoretical bound that decomposes the distilled student model's generalized error into three components: (i) the teacher's generalized error, which acts as a fixed overhead; (ii) the representation distributional discrepancy between the teacher and student models -- \textbf{feature alignment}; and (iii) the prediction distributional discrepancy between the teacher and student models -- \textbf{label alignment}.~To the best of our knowledge, this is the first generalization bound for CMKD that reveals a synergistic influence of teacher quality, feature alignment, and label alignment on the student's generalization (Section~\ref{sec:theory}).

%\textbf{2.~Algorithm Design}.~Based on theoretical analysis, we develop a practical algorithm that enables effective cross-modal knowledge distillation by aligning distributions rather than individual samples via minimizing feature and label alignment, thereby eliminating the need for data-level pairing. Our approach constructs an optimizable surrogate that serves as a medium for selectively distilling teacher knowledge relevant to the student's knowledge. This surrogate is optimized in a bi-level optimization manner by optimizing the student's objective with calibration to feature alignment and label alignment (Section~\ref{sec:algo_design}).

\textbf{2.~Algorithm Design}.~Motivated by our theoretical analysis, we develop a practical framework for cross-modal knowledge distillation that focuses on distribution-level alignment which does not require training examples of sample-level pairing.~In particular, the developed method enables effective knowledge transfer via minimizing both feature and label alignment.~To achieve this, we construct an optimizable surrogate that serves as a medium for selectively distilling teacher knowledge relevant to the student's knowledge.~This surrogate is optimized in a bi-level optimization manner by optimizing the student's objective with calibration to feature alignment and label alignment (Section~\ref{sec:algo_design}).

\textbf{3.~Empirical Evaluation}.~We evaluate our approach on four multimodal benchmarks:~AVE \citep{tian2018audiovisualeventlocalizationunconstrained} for event localization; CREMA-D \citep{cao2014cremed} and RAVDESS \citep{LivingstoneRusso2018} for emotion recognition; and VGGSound \citep{chen2020vggsoundlargescaleaudiovisualdataset}, a large-scale benchmark containing over 200,000 videos spanning more than 300 classes.~Across all datasets, our method consistently outperforms recent state-of-the-art (SOTA) baselines in both unpaired and paired CMKD settings, demonstrating its effectiveness and robustness (Section~\ref{sec:empirical_analysis}).

For interested readers, we also provide a comprehensive literature review of existing KD methods in both in-modal and cross-modal settings in Section~\ref{sec:related}.\vspace{-3mm}

%\textbf{3.~Evaluation}.~We evaluate our approach on four multi-modal benchmarks: AVE \citep{tian2018audiovisualeventlocalizationunconstrained} for event localization; CREMA-D \citep{cao2014cremed} and RAVDESS \citep{LivingstoneRusso2018} for emotion recognition; and VGGsound \citep{chen2020vggsoundlargescaleaudiovisualdataset}, a large-scale dataset featuring over 200,000 videos across 300+ classes.  Across these datasets, our method consistently outperforms recent state-of-the-art baselines in both settings: (1) unpaired data CMKD and (2) paired data CMKD. These results demonstrate the effectiveness and robustness of the proposed approach (Section~\ref{sec:empirical_analysis}).

\section{Theoretical Analysis}
\label{sec:theory}
We begin by formalizing the problem setting and introducing key notations (Section~\ref{sec:notations}). We then establish generalization bounds for the student model under both infinite-data (Section~\ref{sec:thrm_infinity}) and finite-data (Section~\ref{sec:thrm_finite}) settings.\vspace{-2mm}

\subsection{Problem Setting and Notations}
\label{sec:notations}
%Let $\mathcal{M}_T \triangleq (\theta, p_T(y \mid \theta(\boldsymbol{x}^T)))$ denote the teacher model which comprises a feature extractor $\theta(\boldsymbol{x}^T)$ and a solution head $p_T(y\mid \boldsymbol{x}^T)$.~We assume the teacher model was pre-trained using a dataset $ \{ \boldsymbol{x}^T_i, y_i\}_{i=1}^{N_T} \sim D^{T}$ with $(\boldsymbol{x}^T_i, y_i ) \in (\mathcal{X}^T, \mathcal{Y})$.~Our objective is to distill the knowledge from teacher model to the student model $\mathcal{M}_S \triangleq (\phi, p_S(y \mid \phi(\boldsymbol{x}^S)))$, which trained on the student dataset $ \{\boldsymbol{x}^S_j, y_j \}_{j=1}^{N_S} \sim D^{S}$ with $(\boldsymbol{x}^S_j, y_j) \in (\mathcal{X}^S, \mathcal{Y})$.

Let $M_T \triangleq (\theta, p_T(y \mid \boldsymbol{z} = \theta(\boldsymbol{x}^T)))$ denote the teacher model which comprises a feature extractor $\theta(\boldsymbol{x}^T)$ and a prediction head $p_T(y \mid \boldsymbol{z} = \theta(\boldsymbol{x}^T))$.~The teacher model is pre-trained on a dataset $D^T = {(\boldsymbol{x}_i^T, y_i)}_{i=1}^{n_T} \sim (\mathcal{X}^T \times \mathcal{Y})$.

Our goal is to distill relevant knowledge from $M_T$ into a student model $M_S \triangleq (\phi, p_S(y \mid \phi(\boldsymbol{x}^S)))$ with feature extractor $\phi(\boldsymbol{x}^S)$ and solution head $p_S$, so that it can generalize better from a student dataset ${(\boldsymbol{x}_i^S, y_i)}_{i=1}^{n_S} \sim D^S$ with $(\boldsymbol{x}_i^S, y_i) \in \mathcal{X}^S \times \mathcal{Y}$ to unseen data sampled from $(\mathcal{X}^S, \mathcal{Y})$.

We assume the training input to the student model is sampled from new data modalities $\mathcal{X}^S \ne \mathcal{X}^T$ which were not previously seen by the teacher model during its pre-training.~To facilitate representation alignment during distillation, we configure the teacher and student feature extractors, $\theta: \mathcal{X}^T \rightarrow \mathcal{Z}$ and $\phi: \mathcal{X}^S \rightarrow \mathcal{Z}$, to map their respective inputs into a shared embedding space $\mathcal{Z}$. 
\begin{definition}[Generalized Error]
\label{def:err}
We have the generalized error for the teacher model and the student model under the feature maps $\phi$ and $\theta$ as: \vspace{-2mm}
\begin{eqnarray}
\hspace{-8.5mm}\text{err}_{T}(\theta) &\triangleq& \mathbb{E}_{D^T(\boldsymbol{x}^T, y)}\left[- \log p_T\Big(y \mid \theta\left(\boldsymbol{x}^T\right)\Big) \right], \\
\hspace{-8.5mm}\text{err}_{S}(\phi) &\triangleq& \mathbb{E}_{D^S(\boldsymbol{x}^S, y)}\left[-\log p_S\Big(y \mid \phi\left(\boldsymbol{x}^S\right)\Big) \right], \label{CE_loss}
\end{eqnarray}

\vspace{-2mm}
which are the expected (generalized) teacher and student prediction losses over the corresponding data distributions.
\end{definition}

\begin{definition}[Feature Distribution]\label{def:feature_distribution}
The feature distributions $D^S(\boldsymbol{z})$ and $D^T(\boldsymbol{z})$ denote the push-forward distributions induced by the student and teacher feature maps, $\boldsymbol{z} = \phi(\boldsymbol{x}^S)$ and $\boldsymbol{z} = \theta(\boldsymbol{x}^T)$, on the marginal input distributions $D^S(\boldsymbol{x}^S)$ and $D^T(\boldsymbol{x}^T)$, respectively.
\end{definition}%\vspace{-2mm}

Let $D^S(y \mid \boldsymbol{z})$ and $D^T(y \mid \boldsymbol{z})$ denote the student and teacher feature-label conditional distributions induced by the feature maps $\boldsymbol{z} = \phi(\boldsymbol{x}^S)$ and $\boldsymbol{z} = \theta(\boldsymbol{x}^T)$, respectively.

\subsection{Asymptotic Performance Bound}
\label{sec:thrm_infinity}
We will now provide the generalized performance bound for cross-modal distillation in the asymptotic regime where key quantities (see below) in the bound are computed with respect to the true data distributions.~This is equivalent to assuming an infinite amount of data, such that empirical estimates coincide with their population counterparts.

Our main result characterizes the generalized student loss $\mathrm{err}_S(\phi)$ in terms of the following key quantities:\vspace{1mm}

\noindent {\bf Overhead.}~The generalized teacher loss $\mathrm{err}_T(\theta)$.\vspace{1mm}

\noindent {\bf Feature Alignment (FA).}~A function of the distributional distance between the student and teacher representation distributions, $D^S(\boldsymbol{z})$ and $D^T(\boldsymbol{z})$, induced by $\phi$ and $\theta$.\vspace{1mm} 

\noindent {\bf Label Alignment (LA).}~A function of the distributional distance between the student and teacher predictions, which captures the semantic gap between the student predictor $p_S(y \mid \boldsymbol{z})$ and the teacher predictor $p_T(y \mid \boldsymbol{z})$.\vspace{1mm} 

\noindent An informal statement of our result is stated below.

\begin{theorem}[Informal Statement]
\label{thm:1}Given teacher and student feature maps $\theta$ and $\phi$, we have:
\begin{eqnarray}
\hspace{-5mm}\mathrm{err}_S(\phi) \hspace{-1mm}&\leq&\hspace{-1mm} \mathrm{err}_T(\theta)\nonumber\\
\hspace{-1mm}&+&\hspace{-1mm}{\textbf{Feature Alignment}} + {\textbf{Label Alignment}}\ .
\end{eqnarray}
\end{theorem}
This result reveals how the generalization quality of the teacher model and the interplay between \textbf{feature alignment} and \textbf{label alignment} during distillation will influence the generalized student performance.~The teacher generalized performance essentially acts as a fixed overhead.~The remaining terms show that feature alignment alone is insufficient for effective cross-modal distillation, as aggressive representation alignment may inadvertently enlarge the semantic discrepancy between the teacher and student prediction distributions.~In particular, when alignment induces representations that enlarge this predictive gap, the student model may overfit to misaligned semantic structures, ultimately degrading generalization performance.

%This result reveals how the teacher's quality and the interplay between \textbf{feature alignment} and \textbf{label alignment} shape the student performance. The teacher error acts as a fixed overhead, reflecting the influence of the teacher model. The remaining terms suggest that aligning features without careful calibration may inadvertently increase the semantic prediction distributional distance between the teacher and the student model. For instance, when alignment induces representations that enlarge this gap, it can harm the generalization of the student model by steering the prediction head toward overfitting. 

This insight is made precise via the below formal definitions and theorem statement (Theorem~\ref{thm:2}).
\begin{definition}[Feature Alignment]
\label{def:FA}
Let $\Delta$ denote the set of cost metrics $\delta$ on the pre-trained representation space such that the cross-entropy of the teacher prediction,
\begin{eqnarray}
\hspace{-23mm}\ell_\tau(\boldsymbol{z}) &\triangleq& -\mathbb{E}_{D^{T}(y|\boldsymbol{z})}\Big[\log p_T\left(y\mid \boldsymbol{z}\right)\Big] \ ,   
\end{eqnarray}
is $\tau_\delta$-Lipschitz with respect to $\delta$:
\begin{eqnarray}
\hspace{-25.5mm}\big|\ell_\tau(\boldsymbol{z}_1) - \ell_\tau(\boldsymbol{z}_2)\big| &\leq& \tau_\delta \cdot \delta\big(\boldsymbol{z}_1,\boldsymbol{z}_2\big) \ . 
\end{eqnarray}
The feature alignment under the student feature map $\phi$ is 
\begin{eqnarray}
\hspace{-11mm}\textbf{FA}(\phi,\theta) &\triangleq& \min_{\delta \in \Delta} \Big\{\tau_\delta \cdot W_\delta\Big(D^{T}(\boldsymbol{z}), D^{S}(\boldsymbol{z})\Big) \Big\} \ . \label{eq:FA-def}
\end{eqnarray}
where $W_{\delta}$ is Wasserstein-$1$ distance with cost metric $\delta $.
\end{definition}

\begin{definition}[Label Alignment]
\label{def:LA}
Let $\kappa(y, \boldsymbol{z})$ denote the label transport kernel between the teacher and student predictors, defined as $\kappa(y, \boldsymbol{z}) \triangleq D^{T}(y \mid \boldsymbol{z})/D^{S}(y \mid \boldsymbol{z})$.

The label alignment between the teacher predictor $p_T$ and the student predictor $p_S$ is  then defined as
\begin{eqnarray}
\hspace{-4.5mm}\textbf{LA}(p_S, p_T) \hspace{-2mm}&\triangleq&\hspace{-2mm} - \mathbb{E}_{D^S(\boldsymbol{z}, y)} \left[ \log \left(\frac{p_S(y \mid \boldsymbol{z})}{p_T(y \mid \boldsymbol{z})^{\kappa(y, \boldsymbol{z})}} \right) \right]\ .
\end{eqnarray}
\end{definition}

\noindent The formal statement of our result can now be stated below.

\begin{theorem}[Formal Statement]
\label{thm:2}
Plugging the above definition into the informal statement in Theorem~\ref{thm:1},
\begin{eqnarray}
\hspace{-10mm}\mathrm{err}_S(\phi) \hspace{-2mm}&\leq&\hspace{-2mm} \mathrm{err}_T(\theta) \ +\  
{\textbf{FA}}(\phi, \theta) \ +\  {\textbf{LA}}\left(p_S, p_T\right)\ . \label{eq:bound}
\end{eqnarray}
A detailed proof is provided in Appendix~\ref{app:A}.
\end{theorem}

Theorem~\ref{thm:2} formalizes the earlier intuition by providing a principled characterization of the semantic gap in cross-modal knowledge distillation (CMKD) through both feature alignment ({\bf FA}) and label alignment ({\bf LA}).

\subsection{Finite-Data Performance Bound}
\label{sec:thrm_finite}
In this section, we analyze the finite-sample regime, where only $n_S$ student samples and $n_T$ teacher samples are available.~In particular, we revisit the earlier asymptotic bound by replacing its population-level quantities with their empirical counterparts, leading to the following key quantities characterizing the generalized student loss $\text{err}_{S}(\phi)$:

%In this section, we analyze the finite-sample regime, where only $n_S$ student samples and $n_T$ teacher samples are available. Our main result characterizes the generalized student loss $\text{err}_{S}(\phi)$ through the following key quantities:

%\noindent {\bf Overhead.}~The generalized teacher loss $\mathrm{err}_T$.\vspace{1mm}

\noindent {\bf Empirical Label Alignment (LA$_e$).}~We define the empirical label alignment (\textbf{LA}$_e$) as a Monte Carlo estimate of the exact label alignment (\textbf{LA}) computed from $n_S$ samples:
\begin{eqnarray}
\hspace{-4.5mm}\mathbf{LA}_e(p_S, p_T) \hspace{-2mm}&\triangleq&\hspace{-2mm} -\frac{1}{n_S}\sum_{i=1}^{n_S} \log \left(\frac{p_S(y_i \mid \boldsymbol{z}_i)}{p_{T}(y_i \mid \boldsymbol{z}_{i})^{\kappa(y_i, \boldsymbol{z}_i)}} \right)\ . \label{LA_define}
\end{eqnarray}

\noindent{\bf Empirical Feature Alignment (FA$_e$).}~We define the empirical feature alignment (\textbf{FA}$_e$) as an empirical estimate of the population-level feature alignment (\textbf{FA}) computed from $n_S$ student samples and $n_T$ teacher samples:
\begin{eqnarray}
\hspace{-8.5mm}\mathbf{FA}_e(\phi, \theta) &\triangleq& \min_{\delta \in \Delta} \Big\{\tau_{\delta} W_{\delta}\Big(D^{T}_{n_T}(\boldsymbol{z}), D^{S}_{n_S}( \boldsymbol{z}) \Big) \Big\} \ .\label{FA_define}
\end{eqnarray}
\noindent The formal statement of our finite-data result is stated below.

\begin{theorem}[Formal Statement]
\label{thm:3}
Given the teacher and student feature maps $\theta$ and $\phi$, the following holds with the probability at least $1-3\delta$ where $\delta \in (0, 1/3)$: 
\begin{eqnarray}
\hspace{-9mm}\mathrm{err}_S(\phi) \hspace{-2mm}&\leq&\hspace{-2mm} \mathrm{err}_T(\theta) + 
\mathbf{FA}_e\big(\phi, \theta\big) + \mathbf{LA}_e\big(p_S, p_T\big) \nonumber\\
\hspace{-2mm}&+&\hspace{-2mm} \tau_\delta\sqrt{\frac{\log(2/\delta)}{2}}\left(\frac{1}{\sqrt{n_S}} + \frac{1}{\sqrt{n_T}}\right) \nonumber \\
\hspace{-2mm}&+&\hspace{-2mm} O\left(n_S^{-1/s_1} \right) + O\left( n_T^{-1/s_2}\right)\nonumber\\
\hspace{-2mm}&+&\hspace{-2mm}
O\left(2\sqrt{\frac{2d\log(n_S/d) }{n_S}} + \sqrt{\frac{\log(1/\delta)}{2n_S}}\right)\ ,
\end{eqnarray}
where $s_1$ and $s_2$ are any constants larger than the upper Wasserstein dimensions~\citep{weed2017sharpasymptoticfinitesamplerates} of the student and teacher representation distributions, respectively; and $d$ denotes the VC dimension \citep{mohri2012foundations} characterizing the complexity of the student hypothesis class. A detailed proof is provided in Appendix~\ref{app:B}.
\end{theorem}

Intuitively, Theorem~\ref{thm:3} extends the earlier analysis to the practical finite-sample regime, where only a limited number of teacher and student samples are available. Notably, the appearance of the VC dimension reveals a fundamental trade-off between alignment and model complexity: while a more expressive student model may better align with the teacher, increasing its complexity $d$ also enlarges the generalization gap. Consequently, for a fixed number of student samples $n_S$, increasing model capacity does not necessarily improve performance, as the risk of overfitting becomes increasingly dominant, as reflected by the term $O(\sqrt{d/n_S})$.

%Theorem~\ref{thm:3} further provides the general error bound of the student model in the practical scenario, when given a finite number of data points. Interestingly, the inclusion of the VC-dimension highlights a fundamental trade-off: while a more expressive student model may achieve better alignment, its increased complexity $d$ can loosen the generalization bound. This confirms the intuition that, for a fixed sample size $n_S$, increasing model capacity does not monotonically improve performance, as the risk of overfitting becomes dominant, which is captured by the term $O(\sqrt{d/n_S})$.

\section{Algorithm Design}
\label{sec:algo_design}
\begin{figure*}[t]
    \centering
    \includegraphics[width=\textwidth]{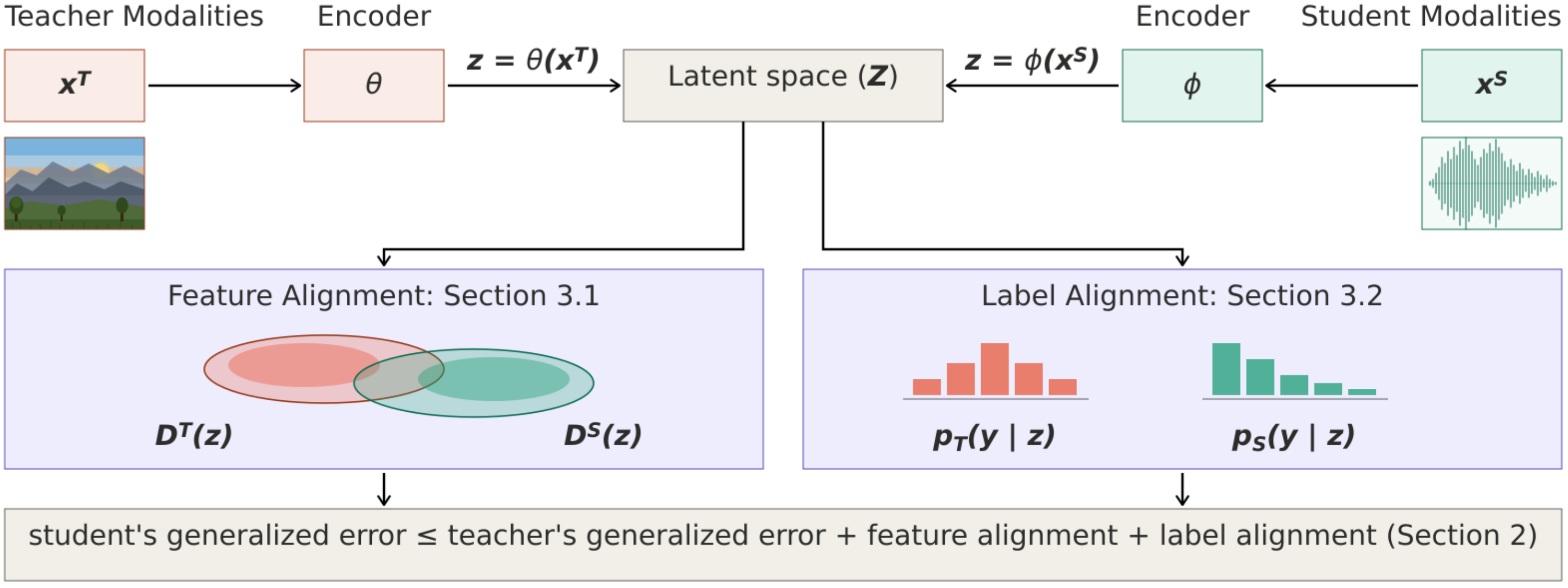}
    \caption{\textbf{Overview of our UCMKD framework}:~The teacher and student encoders map inputs from different modalities into a shared latent space $\mathcal{Z}$.~The cross-modal generalization bound decomposes into two distributional quantities: {\bf Feature Alignment}, a Wasserstein distance between the latent distributions $\mathcal{D}^T(\boldsymbol{z})$ and $\mathcal{D}^S(\boldsymbol{z})$ -- Section~\ref{sec:feature_alignment}; and {\bf Label Alignment}, a distance measure between the induced predictive distributions $p_T(y\mid\boldsymbol{z})$ and $p_S(y\mid\boldsymbol{z})$ -- Section~\ref{sec:label_alignment}.~Theorems~\ref{thm:2} and~\ref{thm:3} bound the student's generalized error by the sum of teacher error, feature alignment, and label alignment, motivating distribution-level alignment without sample-level pairing.}\vspace{-4mm}
    %\caption{\textbf{Overview of our UCMKD framework}:~(a) Overall meta-learning-based optimization of the student objective, where the student model undergoes inner adaptation steps to align with the teacher’s knowledge;~(b) detailed view of the inner adaptation process, decomposed into two stages: stage $1$ ({\bf Feature Alignment}) minimizes the FA loss $\ell_{\mathbf{LA}}$ to align latent representations, while stage $2$ ({\bf Label Alignment}) minimizes the LA loss $\ell_{\mathbf{LA}}$ to align predictive distributions, both with respect to a fixed teacher model.}
    \label{fig:overview}
\end{figure*}

We now present our proposed Cross-Modal Knowledge Distillation framework, \textbf{UCMKD} (\textbf{U}niversal \textbf{C}ross \textbf{M}odal \textbf{K}nowledge \textbf{D}istillation), designed to address the unpaired-data challenge based on the theoretical insights developed in Section~\ref{sec:theory}. Our framework adopts the bi-level optimization approach in~\citep{finn2017modelagnosticmetalearningfastadaptation} that minimizes the generalized student loss (outer optimization) while calibrating the cross-modal semantic gap between teacher and student models (inner optimization).~As suggested by our theoretical analysis, this semantic gap is characterized via two complementary quantities: feature alignment (\textbf{FA}) and label alignment (\textbf{LA}).~Accordingly, UCMKD operates through the two-stage workflow illustrated in Figure~\ref{fig:overview}(b).~In stage $1$, we learn the student encoder $\phi$ by minimizing the representation distributional discrepancy between teacher and student models in the latent space (Section~\ref{sec:feature_alignment}).~In stage $2$, we optimize both the encoder $\phi$ and prediction head $p_S(y \mid \boldsymbol{z})$ to minimize the prediction distributional discrepancy between teacher and student models (Section~\ref{sec:label_alignment}).

%We will now detail our developed Cross Modal Knowledge Distillation framework named \textbf{UCMKD} (\textbf{U}niversal \textbf{C}ross \textbf{M}odal \textbf{K}nowledge \textbf{D}istillation), which addresses the unpaired data challenge based on our theoretical analysis (Section~\ref{sec:theory}).~We use bi-level optimization \citep{finn2017modelagnosticmetalearningfastadaptation} to minimize the generalized loss of the student model (outer update) with calibration to the cross-modal semantic gap between the teacher and the student model (inner update).~This cross-modal gap can be decomposed into feature alignment (\textbf{FA}) and label alignment (\textbf{LA}), which are optimized via a two-stage workflow as illustrated in Figure~\ref{fig:overview}(b).~Stage 1 learns the student encoder $\phi$ via minimizing the representation distributional distance between student and teacher in the latent space (Section~\ref{sec:feature_alignment}).~Stage 2 optimizes both encoder $\phi$ and prediction head $p_S(y \mid \boldsymbol{z})$ to minimize the prediction distributional distance between the teacher and the student model (Section~\ref{sec:label_alignment}).

\noindent To elaborate on this design, we note that directly optimizing the student objective together with the semantic cross-modal gap from Theorem~\ref{thm:2}, as in conventional CMKD approaches, is often unstable due to the highly coupled interaction between the prediction map $p_S(y \mid \boldsymbol{z})$ and the feature map $\boldsymbol{z} = \phi(\boldsymbol{x}^{S})$ (see Table~\ref{tab:component_contribution_ablation}).~In particular, update to the feature extractor $\phi$ simultaneously modifies the alignment between teacher and student distributions while reshaping the student representation landscape $D^S(\boldsymbol{z})$ on which the predictor $p_S(y \mid \boldsymbol{z})$ is learned.~This moving-target effect substantially complicates the optimization landscape and can lead to unstable convergence.~To address this challenge, we adopt a hybrid bi-level optimization approach with a structured two-stage inner-update procedure that calibrates the student loss with respect to the cross-modal semantic gap.~In particular, the inner optimization is decomposed into minimizing feature alignment loss (stage $1$) and minimizing label alignment loss (stage $2$).~Figure~\ref{fig:overview} provides an overview of the proposed framework, which optimizes the student loss via inner adaptation steps used to compute the outer optimization loss in a meta-learning fashion.~The pseudo-code of the full algorithm is provided in Algorithm~\ref{algo:metaKD}.
 
%\noindent To elaborate on this design, we note that a direct minimization of the jointly student objective and the semantic cross modal gap (Theorem~\ref{thm:2}) as traditional CMKD is, however, unstable due to the entangled effect of optimizing both the target prediction map $p_S(y \mid \boldsymbol{z})$ and the feature map $\boldsymbol{z} = \phi(\boldsymbol{x}^{S})$ in complex and interdependent ways (See Table~\ref{tab:component_contribution_ablation}).~In particular, changes in the representation $\phi$ simultaneously alter the alignment between the student and teacher distributions reshape the student feature landscape $D^S(\boldsymbol{z})$, which predictor $p_S(y \mid \boldsymbol{z})$ is optimized on. This moving target phenomenon complicates the optimization landscape and often leads to destabilized convergence. To mitigate this, we adopt a hybrid bi-level optimization featuring a structured two-stage inner update approach that optimizes the objective of the student model with calibration to the modality gap, which is decomposed into minimizing feature alignment loss (Stage 1) and minimizing label alignment loss (Stage 2). Figure~\ref{fig:overview} shows the overview of our proposed method, which minimizes the objective loss in meta-learning style via virtual update. The pseudo-code of our algorithm is provided in Algorithm~\ref{algo:metaKD}.
\begin{algorithm}[t]
\caption{Universal Cross-Modal KD (\textbf{UCMKD})}
\begin{algorithmic}[1]
\label{algo:metaKD}
\STATE \textbf{input:}~teacher model $M_{T} \triangleq (\theta, p_T( y\mid \boldsymbol{z} = \theta(\boldsymbol{x}^T)))$, student and teacher datasets: $D^S$ and $D^T$,\\ numbers of outer update epochs $n_0$,\\ inner adaptation epochs ($n_1$ and $n_2$),\\ learning rate $\eta$, \\ and hyper-parameters $\lambda_1, \lambda_2$.
\STATE \textbf{output:}~student model $M_{S} \triangleq (\phi, p_S( y\mid \boldsymbol{z} = \phi(\boldsymbol{x}^S)))$
\STATE initialize the student encoder and predictor $\phi^{0}, p^{0}_S$.
\FOR{$t = 1$ to $n_0$}
    \STATE $\phi_{\mathrm{tmp}} \leftarrow \phi^{t-1}$ and $p_{\mathrm{tmp}} \leftarrow p^{t-1}_S$

     \FOR{$r = 1$ to $n_1$}
        \STATE $\phi_{\mathrm{tmp}} \leftarrow 
        \phi_{\mathrm{tmp}} - \eta \lambda_1 
        \nabla_{\phi} \ell_{\mathbf{FA}}\big(\phi_{\mathrm{tmp}}\big)$
        \hfill \# Eq.~\eqref{FA_loss}
    \ENDFOR

   \FOR{$r = 1$ to $n_2$} %\hfill \# Eq.~\eqref{LA_loss}
        \STATE \# see Eq.~\eqref{LA_loss} 
        \STATE $\phi_{\mathrm{tmp}} \leftarrow
        \phi_{\mathrm{tmp}} -\lambda_2 \eta \nabla_{\phi} \ell_{\mathbf{LA}}\big(\phi_{\mathrm{tmp}}, p_{\mathrm{tmp}}\big)$
        
        \STATE $p_{\mathrm{tmp}} \leftarrow
        p_{\mathrm{tmp}} -\lambda_2 \eta \nabla_{p} \ell_{\mathbf{LA}}\big(\phi_{\mathrm{tmp}}, p_{\mathrm{tmp}}\big)$

    \ENDFOR
    
    \STATE $\phi^{t} \ =\  \phi^{t-1} \ -\ \eta \nabla_{\phi} \mathrm{err}_S\big(\phi_{\mathrm{tmp}}, p_{\mathrm{tmp}}\big)$ \hfill \# Eq.~\eqref{CE_loss}
    \STATE $p^{t}_S \ =\ p^{t-1}_{S} \ -\  \eta \nabla_{p} \mathrm{err}_S\big(\phi_{\mathrm{tmp}}, p_{\mathrm{tmp}}\big)$ \hfill \# Eq.~\eqref{CE_loss}
\ENDFOR

\STATE \textbf{return} distilled student model $M_S = (\phi^{n_0}, p^{n_0}_S)$%\vspace{-20mm}
\end{algorithmic}
\end{algorithm}%\vspace{-20mm}

\subsection{Feature Alignment Loss}
\label{sec:feature_alignment}
To compute the feature alignment (\textbf{FA}), a direct approach is to minimize the optimal transport distance between the empirical student and teacher representation distributions in the shared latent space $\mathcal{Z}$:
%The objective of feature alignment (\textbf{FA}) is to align the teacher's and student's representation in the shared latent space $\mathcal{Z}$. Suppose that we have the optimal transportation cost between the student and the teacher representation distribution, which is provided by the $n$ and $m$ numbers of samples, respectively: 
\begin{eqnarray}
\hspace{-6mm}W_{\delta}\Big(D^S_{n_S}(\boldsymbol{z}), D^T_{n_T}(\boldsymbol{z})\Big) \hspace{-2.5mm}&=&\hspace{-2.5mm}\min_{\pi \in \Pi} \sum_{i=1}^{n_S} \sum_{j=1}^{n_T} \pi_{ij} \delta\Big(\boldsymbol{z}^{S}_i, \boldsymbol{z}^{T}_j\Big)
\end{eqnarray}
where $\Pi$ denote the set of coupling $\pi$ between $D_{n_S}^S(\boldsymbol{z})$ and $D_{n_T}^T(\boldsymbol{z})$\footnote{The corresponding marginals of $\pi(\boldsymbol{z}^S_i, \boldsymbol{z}^T_j)$ over $\boldsymbol{z}_j^T$ and $\boldsymbol{z}_i^S$ correspond to $D_{n_S}^S(\boldsymbol{z}_i^S)$ and $D_{n_T}^T(\boldsymbol{z}_j^T)$.} and $\pi_{ij} \triangleq \pi(\boldsymbol{z}_i^S, \boldsymbol{z}_j^T)$.~This can be naively achieved via solving a linear program with a computational complexity $O(\max(n_S, n_T)^3)$ \citep{peyré2020computationaloptimaltransport}.

To reduce this complexity, we instead adopt the following entropic-regularized formulation:
\begin{eqnarray}
\hspace{-7.5mm}\pi^* \hspace{-2mm}&=&\hspace{-2mm} \arg \min_{\pi \in \Pi} \left(\sum_{i=1}^{n_S} \sum_{j=1}^{n_T} \pi_{ij} \delta\Big(\boldsymbol{z}^{S}_i, \boldsymbol{z}^{T}_j\Big) + \epsilon H(\pi)\right).
\end{eqnarray}
This regularized problem can be solved efficiently using the Sinkhorn algorithm with quadratic complexity~\citep{peyré2020computationaloptimaltransport}.~In practice, we use the Euclidean transport cost $\delta \triangleq \ell_2$ and the default regularization parameter $\epsilon=0.1$. The resulting feature alignment loss is defined as
%Such entropic-regularization form can be solved efficiently using the Sinkhorn algorithm, with overall complexity of \(O(n^2)\), which offers significant speedup for large-scale problems~\citep{peyré2020computationaloptimaltransport}.~For a practical implementation, we use the default regularization parameter \(\epsilon = 0.1\) and the Euclidean cost metric $\delta \triangleq \ell_2$ to ensure a balance between computational efficiency and accuracy. To simplify, we treat the weight of feature alignment as the hyperparameter $\lambda_1$ (see Algorithm~\ref{algo:metaKD}), thus we only compute the Feature Alignment loss as the Wasserstein distance with $N_S$ student data points and $N_T$ teacher data points:
\begin{eqnarray}
\hspace{-21mm}\ell_{\mathbf{FA}}(\theta,\phi) &\triangleq& W_{\ell_{2}}\Big(D^S_{N_S}(\boldsymbol{z}), D^{T}_{N_T}(\boldsymbol{z}) \Big) \ , \label{FA_loss}
\end{eqnarray}
where the contribution of feature alignment is controlled by the hyperparameter $\lambda_1$ (Algorithm~\ref{algo:metaKD}).~Here, $D_{n_S}^S(\boldsymbol{z})$ and $D_{n_T}^T(\boldsymbol{z})$ denote the push-forward of the empirical student and teacher input distributions under the corresponding feature maps $\phi$ and $\theta$, respectively.  

\subsection{Label Alignment Loss}
\label{sec:label_alignment}
%The objective of Label Alignment (LA) is to synchronize the predictive distributions of the teacher and student models given an identical latent embedding $\boldsymbol{z} \in \mathcal{Z}$. While aligning predictions is a cornerstone of Knowledge Distillation, our formulation introduces a label transport kernel, $\kappa(y, \boldsymbol{z})$, to adaptively weight the distillation process. Given $N_S$ student samples, the LA loss is defined as:
The label alignment loss is set to be the empirical {\bf LA}$_e$ term previously defined in Eq.~\eqref{LA_define}: 
\begin{eqnarray}
\hspace{-7.5mm}\ell_{\mathbf{LA}}\big(p_S, p_T\big)  \hspace{-2mm}&\triangleq&\hspace{-2mm} -\frac{1}{n_S}\sum_{i=1}^{n_S} \log \left( \frac{p_S(y_i \mid \boldsymbol{z}_i)}{p_{T}(y_i \mid \boldsymbol{z}_{i})^{\kappa(y_i, \boldsymbol{z}_i)}} \right) \label{LA_loss}
\end{eqnarray}
where $\kappa(y, \boldsymbol{z}) \triangleq D^T(y\mid\boldsymbol{z})/D^S(y\mid \boldsymbol{z})$.

Intuitively, $\kappa(y, \boldsymbol{z})$ quantifies the agreement between teacher and student prediction distributions, thereby acting as a gating mechanism for \textbf{selective knowledge distillation}.~A practical estimation procedure for the transport kernel is provided in Appendix~\ref{sec:estimate_kernel}.~When the teacher’s prediction conflicts with the student’s target semantics, particularly when the teacher assigns negligible probability to the target label, the kernel approaches zero, $\kappa(y, \boldsymbol{z}) \simeq 0$.~In this regime, the alignment loss naturally reduces to the standard supervised loss of the student model:
\begin{eqnarray}
\hspace{-9mm}\lim_{\kappa \to 0} \ell_{\mathbf{LA}} \hspace{-2mm}&=&\hspace{-2mm} \mathbb{E}_{D^S(\boldsymbol{z},y)} \left[ -\log p_S(y \mid \boldsymbol{z}) \right] \ =\  \text{err}_{S}(\phi).
\end{eqnarray}
This mechanism ensures that, under semantic disagreement, the student model prioritizes its own supervised signal rather than inheriting potentially unreliable teacher guidance.~Furthermore, the formulation naturally enables a pseudo-labeling strategy~\citep{nguyen2020leepnewmeasureevaluate} for minimizing the semantic prediction discrepancy between teacher and student models without requiring paired data, thus improving robustness in unpaired settings.

\iffalse
The kernel $\kappa(y, \boldsymbol{z})$ quantifies the consensus between the teacher and student distributions, acting as a gating mechanism for \textbf{Selective Knowledge Distillation}. The practical estimation for the transport kernel is provided in Appendix~\ref{sec:estimate_kernel}. In scenarios where the teacher’s knowledge conflicts with the student’s knowledge, specifically when the teacher assigns a negligible probability to the target label, the kernel vanishes ($\kappa(y, \boldsymbol{z}) \approx 0$). In this limit, the alignment objective reduces to the standard empirical risk of the student:
\begin{equation}
\lim_{\kappa(y,\boldsymbol{z}) \to 0} \mathcal{L}_{LA} = \mathbb{E}_{D^S(\boldsymbol{z},y)} \left[ -\log p_S(y \mid \boldsymbol{z}) \right] = \text{err}_{S}
\end{equation}
Crucially, this mechanism ensures that during knowledge conflict, the student prioritizes its own supervised signal over the potentially unreliable teacher guidance. Furthermore, this formulation enables a pseudo-labeling strategy~\citep{nguyen2020leepnewmeasureevaluate} that minimizes the semantic prediction distributional gap between models without requiring strictly paired data, enhancing the method's versatility across unpaired settings. 
\fi

\section{Empirical Evaluation}
\label{sec:empirical_analysis}
\begin{table*}[t]
\centering
\caption{Prediction accuracy on the {\bf AVE}, {\bf RAVDESS}, {\bf CREMA-D}, and {\bf VGGSound} datasets under the unpaired setting, comparing {\bf Cross-Entropy}, {\bf Feature-Based Knowledge Distillation}, and our proposed {\bf UCMKD} framework.~For reference, we also report {\bf Vanilla KD}~\citep{hinton2015distillingknowledgeneuralnetwork} trained under paired supervision.~$A \to V$ and $V \to A$ denote distillation from audio to visual and visual to audio modalities, respectively.~Despite operating without paired supervision, {\bf UCMKD} consistently outperforms unpaired baselines and surpasses the paired {\bf Vanilla KD} baseline on 6 out of 8 experimental tasks.}
\label{tab:unpaired_results}
\small
\setlength{\tabcolsep}{4pt}

\resizebox{\linewidth}{!}{
\begin{tabular}{|l|c|c|c|c|c|c|c|c|}
\hline
\bf Method 
& \multicolumn{2}{c|}{\bf AVE} 
& \multicolumn{2}{c|}{\bf RAVDESS} 
& \multicolumn{2}{c|}{\bf CREMA-D} 
& \multicolumn{2}{c|}{\bf VGGsound} \\
\cline{2-9}
& $A\!\to\!V$ & $V\!\to\!A$
& $A\!\to\!V$ & $V\!\to\!A$
& $A\!\to\!V$ & $V\!\to\!A$
& $A\!\to\!V$ & $V\!\to\!A$ \\
\hline
Teacher
& 52.74 & 30.35
& 79.92 & 77.72
& 65.46 & 70.97
& 56.78 & 44.43 \\
\hline
Cross Entropy
& 27.70 $\pm$ 2.35 & 50.08 $\pm$ 2.88
& 65.47 $\pm$ 3.69 & 70.66 $\pm$ 1.16
& 71.51 $\pm$ 0.88 & 61.96 $\pm$ 0.83
& 41.68 $\pm$ 2.32 & 54.40 $\pm$ 2.34 \\

Feature KD
& 31.01 $\pm$ 1.04 & 48.51 $\pm$ 1.49
& 65.37 $\pm$ 3.20 & 69.80 $\pm$ 2.96
& 69.22 $\pm$ 1.38 & 61.69 $\pm$ 0.69
& 41.07 $\pm$ 1.84 & 52.08 $\pm$ 0.83 \\
\hline
Vanilla KD
& 29.85 $\pm$ 1.85 & 49.01 $\pm$ 2.47
& 67.17 $\pm$ 3.86 & 73.10 $\pm$ 1.63
& \textbf{72.18 $\pm$ 1.15} & 62.45 $\pm$ 0.56
& \textbf{43.35 $\pm$ 0.29} & 51.71 $\pm$ 0.69 \\
\hline
\textbf{UCMKD}
& \textbf{34.16 $\pm$ 1.12} & \textbf{52.24 $\pm$ 1.08}
& \textbf{73.83 $\pm$ 1.25} & \textbf{74.43 $\pm$ 2.15}
& 71.64 $\pm$ 0.86 & \textbf{66.67 $\pm$ 1.24}
& 43.10 $\pm$ 0.38 & \textbf{56.84 $\pm$ 0.47} \\
\hline
\end{tabular}
}

\iffalse
\begin{tabular}{l *{8}{c}}
\toprule
Method & \multicolumn{2}{c}{AVE} & \multicolumn{2}{c}{RAVDESS} & \multicolumn{2}{c}{CREMA-D} & \multicolumn{2}{c}{VGGsound} \\
\cmidrule(lr){2-3} \cmidrule(lr){4-5} \cmidrule(lr){6-7} \cmidrule(lr){8-9}
       & $A\!\to\!V$ & $V\!\to\!A$ & $A\!\to\!V$ & $V\!\to\!A$ & $A\!\to\!V$ & $V\!\to\!A$ & $A\!\to\!V$ & $V\!\to\!A$ \\
\midrule
Teacher           & 52.74 & 30.35 & 79.92 & 77.72 & 65.46 & 70.97 & 56.78 & 44.43 \\
\midrule
CE     & 27.70  & 50.08 & 65.47  & 70.66  & 71.51  & 61.96  & 41.68  & 54.40  \\
Feature KD   & 31.01  & 48.51  & 65.37  & 69.80  & 69.22 & 61.69  &  41.84 & 52.08  \\
\midrule
Vanilla KD       & 29.85  & 49.01  & 67.17  & 73.10  & \textbf{72.18} & 62.45   & \textbf{43.34}  & 51.71 \\
\midrule
\textbf{UCMKD}     & \textbf{34.16 } & \textbf{52.24} & \textbf{73.83 } & \textbf{74.43 } & 71.64 & \textbf{66.67} & 43.10  & \textbf{56.84 } \\
\bottomrule
\end{tabular}
\fi

\end{table*}

In this section, we present comprehensive experimental results validating the effectiveness of our method under both unpaired-data (Section~\ref{sec:unpaired_results}) and paired-data (Section~\ref{sec:paired_results}) settings.~We further provide ablation studies analyzing parameter sensitivity and performance under data-scarcity scenarios (Section~\ref{sec:ablation_results}).~Due to limited space, additional experimental results and ablation studies are deferred to Appendix~\ref{sec:addtional_exp_results}.

\begin{table*}[t]
\centering
\caption{Prediction accuracy on the {\bf AVE}, {\bf RAVDESS}, {\bf CREMA-D}, and {\bf VGGSound} datasets under the paired setting, comparing {\bf Vanilla KD}~\citep{hinton2015distillingknowledgeneuralnetwork}, {\bf C2KD}~\citep{Huo2024C2KD}, {\bf DKD}~\citep{Zhao2022DKD}, {\bf RKD}~\citep{park2019relationalknowledgedistillation}, {\bf RLD}~\citep{sun2025knowledgedistillationrefinedlogits}, {\bf FitNet}~\citep{romero2015fitnetshintsdeepnets}, {\bf Review}~\citep{Chen2021Review}, and our {\bf UCMKD} framework.~$A \to V$ and $V \to A$ denote distillation from audio to visual and visual to audio modalities, respectively.~{\bf UCMKD} achieves the best performance on 5 out of 8 experimental tasks.}
\label{tab:paired_results}
\small
\setlength{\tabcolsep}{6pt}

\resizebox{\linewidth}{!}{
\begin{tabular}{|l|c|c|c|c|c|c|c|c|}
\hline
\bf Method 
& \multicolumn{2}{c|}{\bf AVE} 
& \multicolumn{2}{c|}{\bf RAVDESS} 
& \multicolumn{2}{c|}{\bf CREMA-D} 
& \multicolumn{2}{c|}{\bf VGGsound} \\
\cline{2-9}
& $A\!\to\!V$ & $V\!\to\!A$
& $A\!\to\!V$ & $V\!\to\!A$
& $A\!\to\!V$ & $V\!\to\!A$
& $A\!\to\!V$ & $V\!\to\!A$ \\
\hline

Vanilla KD
& 29.85 $\pm$ 1.85 & 49.01 $\pm$ 2.47
& 67.17 $\pm$ 3.86 & 73.10 $\pm$ 1.63
& \textbf{72.18 $\pm$ 1.15} & 62.45 $\pm$ 0.56
& 43.35 $\pm$ 0.29 & 51.71 $\pm$ 0.69 \\

RLD
& 22.80 $\pm$ 1.22 & 42.87 $\pm$ 0.82
& 56.64 $\pm$ 1.98 & 63.94 $\pm$ 0.85
& 43.54 $\pm$ 4.59 & 53.04 $\pm$ 0.67
& 32.73 $\pm$ 0.66 & 44.66 $\pm$ 0.70 \\

RKD
& 27.86 $\pm$ 0.70 & 42.54 $\pm$ 0.54
& 41.60 $\pm$ 6.33 & 39.50 $\pm$ 1.53
& 44.44 $\pm$ 2.79 & 62.50 $\pm$ 0.87
& 37.30 $\pm$ 0.49 & 50.71 $\pm$ 0.40 \\

DKD
& 22.80 $\pm$ 0.65 & 34.08 $\pm$ 1.66
& 62.67 $\pm$ 4.27 & 62.27 $\pm$ 2.57
& 30.02 $\pm$ 6.11 & 57.40 $\pm$ 0.38
& 35.70 $\pm$ 0.28 & 43.85 $\pm$ 0.29 \\

C2KD
& 33.33 $\pm$ 0.73 & 47.15 $\pm$ 1.61
& 56.41 $\pm$ 2.42 & \textbf{82.78 $\pm$ 0.41}
& 71.50 $\pm$ 0.11 & 64.43 $\pm$ 0.42
& 40.90 $\pm$ 0.30 & \textbf{61.90 $\pm$ 0.27} \\

FitNet
& 25.87 $\pm$ 1.95 & 49.25 $\pm$ 1.61
& 68.08 $\pm$ 0.75 & 69.96 $\pm$ 3.43
& 70.11 $\pm$ 1.32 & 65.01 $\pm$ 0.01
& 37.90 $\pm$ 0.39 & 57.10 $\pm$ 0.79 \\

Review
& 22.30 $\pm$ 0.62 & 48.92 $\pm$ 0.65
& 54.91 $\pm$ 3.20 & 71.50 $\pm$ 2.00
& 63.89 $\pm$ 1.68 & 61.02 $\pm$ 0.54
& 38.20 $\pm$ 0.47 & 57.90 $\pm$ 0.79 \\
\hline

\textbf{UCMKD}
& \textbf{33.50 $\pm$ 1.82} & \textbf{53.07 $\pm$ 0.51}
& \textbf{76.06 $\pm$ 2.28} & 75.13 $\pm$ 0.65
& 70.43 $\pm$ 0.66 & \textbf{66.75 $\pm$ 1.36}
& \textbf{43.70 $\pm$ 0.182} & 55.98 $\pm$ 0.38 \\
\hline
\end{tabular}
}

\iffalse
\begin{tabular}{l *{8}{c}}
\toprule
Method & \multicolumn{2}{c}{AVE} & \multicolumn{2}{c}{RAVDESS} & \multicolumn{2}{c}{CREMA-D} & \multicolumn{2}{c}{VGGsound} \\
\cmidrule(lr){2-3} \cmidrule(lr){4-5} \cmidrule(lr){6-7} \cmidrule(lr){8-9}
       & $A\!\to\!V$ & $V\!\to\!A$ & $A\!\to\!V$ & $V\!\to\!A$ & $A\!\to\!V$ & $V\!\to\!A$  & $A\!\to\!V$ & $V\!\to\!A$\\
\midrule

Vanilla KD        & 29.85  & 49.01  & 67.17  & 73.10  & \textbf{72.18} & 62.45 & 43.30 &  51.70\\
RLD & 22.80  & 42.87  & 56.64  & 63.94  & 43.54  & 53.04  & 32.73 & 44.65  \\ 
RKD & 27.86  & 42.54  & 41.60  & 39.50  & 44.44  & 62.50 & 37.30 & 50.71  \\
DKD & 22.80  & 34.08  & 62.67  & 62.27  & 30.02  & 57.40  & 35.70 & 43.85\\ 
C2KD & 33.33  & 47.15  & 56.41  & \textbf{82.78 } & 71.50  & 64.43 & 40.90 & \textbf{61.90}  \\
FitNet & 25.87 & 49.25 & 68.01 & 69.96 & 70.11 & 65.00 & 37.93 & 57.11 \\
Review & 23.30 & 48.92 & 54.91 & 71.50 & 63.89 & 61.02 & 38.20 & 57.90 \\
\midrule
\textbf{UCMKD}     & \textbf{33.50} & \textbf{53.07} & \textbf{76.06} & 75.13  & 70.43  & \textbf{66.75} & \textbf{43.70} & 55.98\\
\bottomrule
\end{tabular}
\fi
\end{table*}

\subsection{Implementation Details}
We evaluate our proposed cross-modal distillation method {\bf UCMKD} on $4$ multi-modal datasets which include: (1) \textbf{AVE}~\citep{tian2018audiovisualeventlocalizationunconstrained} is an audio-visual dataset for audio-visual event localization, which has 28 classes; (2) \textbf{CREMA-D}~\citep{cao2014cremed} is an audio-visual dataset for speech emotion recognition,
with 6 categorizations; (3) \textbf{RAVDESS}~\citep{LivingstoneRusso2018} is an audio-visual dataset containing 1,440 emotional utterances with 8 different emotion classes; (4) \textbf{VGGsound}~\citep{chen2020vggsoundlargescaleaudiovisualdataset} is a large-scale video dataset containing about 200 000 videos and more than 300 classes covering daily life activities. 

\textbf{Unpaired Data Simulation.}~To simulate the unpaired setting, we apply a stochastic permutation to the original multimodal dataset $\{\mathcal{X}_1, \mathcal{X}_2, \mathcal{Y}\}$.~Specifically, we break the instance-level correspondence between modalities by randomly shuffling the indices of one modality relative to the other, resulting in two independent subsets, $\{\mathcal{X}_1, \mathcal{Y}\}$ and $\{\mathcal{X}_2, \mathcal{Y}\}$.~This procedure removes sample-level alignment while preserving the marginal distributions of each modality.

\textbf{Hyperparameters.}~We use the same hyperparameter configuration across all baselines and train each network for 100 epochs with an initial learning rate of $1\mathrm{e}{-2}$. Following~\citep{peng2022balancedmultimodallearningonthefly}, we adopt ResNet-18 \citep{he2015deepresiduallearningimage} as the backbone architecture for both visual and audio modalities. Additional implementation and hyperparameter details are provided in Appendix~\ref{sec:hyperparams_deatail}.

\subsection{Experiment Results on Unpaired Setting}
\label{sec:unpaired_results}
Given the limited literature on knowledge distillation under unpaired settings, we compare our method ({\bf UCMKD}) against:~(1) \textbf{Cross-Entropy}, which serves as a non-distillation baseline; and (2) \textbf{Feature Distillation}, which performs distillation through latent-space representation alignment.~Table~\ref{tab:unpaired_results} reports the prediction accuracy of all methods across the evaluated tasks.~{\bf UCMKD} consistently achieves the best performance, with an average improvement of approximately $14.3\%$ over {\bf Cross-Entropy} and $7.5\%$ over {\bf Feature Distillation}.~Notably, {\bf UCMKD} outperforms \textbf{Vanilla KD} \citep{hinton2015distillingknowledgeneuralnetwork} on 6 out of 8 tasks, despite {\bf Vanilla KD} benefiting from paired multimodal data, an advantage unavailable to {\bf UCMKD} in our unpaired setting.~These results demonstrate the effectiveness of our method in transferring cross-modal knowledge without requiring explicit sample-level correspondence.

\subsection{Experiment Results on Paired Setting}
\label{sec:paired_results}
In this section, we evaluate {\bf UCMKD} under the paired-data knowledge distillation (KD) setting and compare it against {\bf Vanilla KD}~\citep{hinton2015distillingknowledgeneuralnetwork} and several state-of-the-art baselines, including {\bf C2KD}~\citep{Huo2024C2KD}, {\bf DKD}~\citep{Zhao2022DKD}, {\bf RKD}~\citep{park2019relationalknowledgedistillation}, {\bf RLD}~\citep{sun2025knowledgedistillationrefinedlogits}, {\bf FitNet}~\citep{romero2015fitnetshintsdeepnets}, and {\bf Review}~ \citep{Chen2021Review}.~As shown in Table~\ref{tab:paired_results}, our method consistently outperforms competing baselines on the {\bf AVE}, {\bf RAVDESS}, and {\bf CREMA-D} datasets.~In particular, our framework achieves the highest accuracy on 5 out of 8 experimental tasks.~This is consistent with the unpaired-setting results in Section~\ref{sec:unpaired_results} and further demonstrate the universally robustness of the proposed framework across both paired- and unpaired-data cross-modal distillation settings.

\begin{figure}
    \centering
    \includegraphics[width=1.0\linewidth]{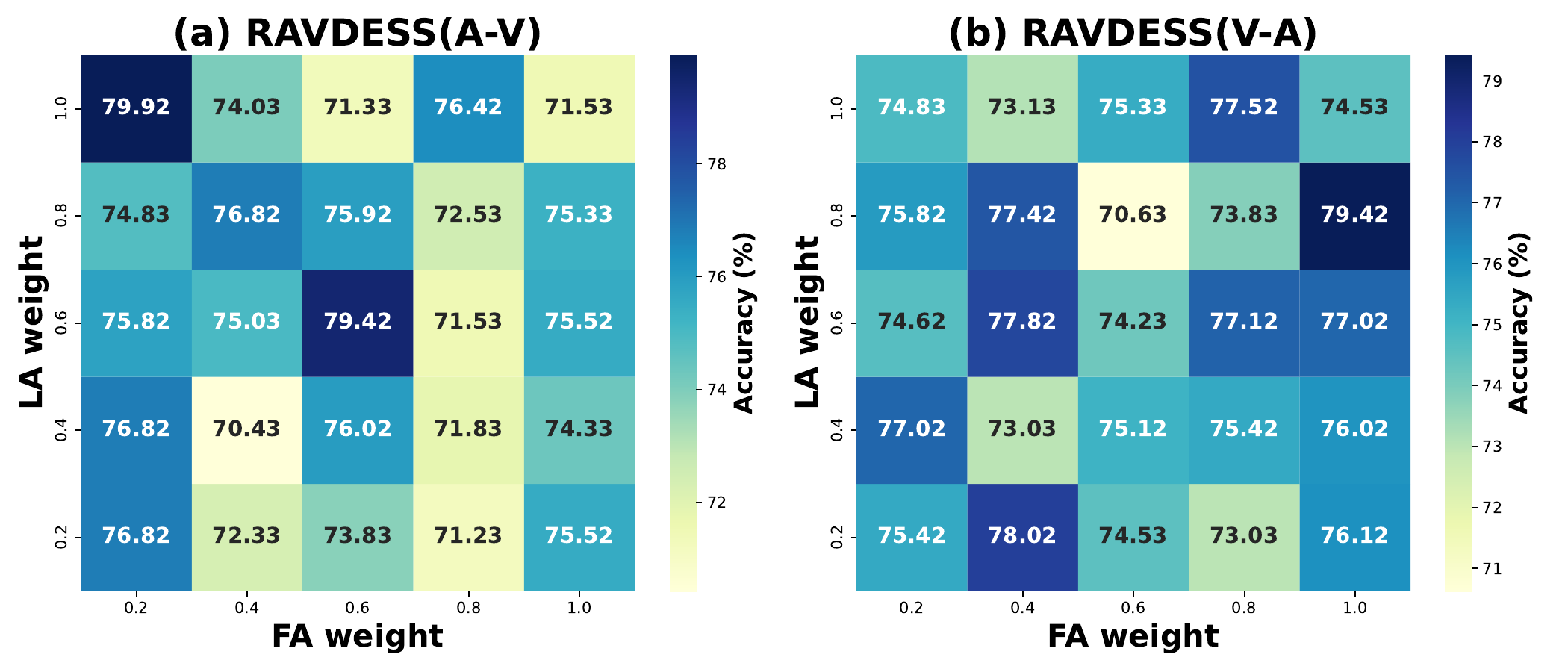}
    \caption{Heatmap of the performance of our method ({\bf UCMKD}) on RAVDESS~\citep{LivingstoneRusso2018} across different values of the hyper-parameters $\lambda_1$ and $\lambda_2$ (Algorithm~\ref{algo:metaKD}) under (a) audio-to-visual ($A \to V$) and (b) visual-to-audio ($V\to A$) settings.}
    \label{fig:heat_map_lamda}
\end{figure}
\subsection{Ablation Studies}
\label{sec:ablation_results}

\begin{table}[h]
\raggedright
\caption{Prediction accuracy on the {\bf RAVDESS} dataset using the ResNet-18 backbone with a training subset ratio of $0.5$, comparing our method {\bf UCMKD} against the {\bf Cross-Entropy}, {\bf Feature-Based Knowledge Distillation}, and {\bf Vanilla KD} baselines.}
\label{tab:ravdess_05}
\begin{tabular}{|lcc|}
\hline
\textbf{Method} & \textbf{$A\!\to\!V$} & \textbf{$V\!\to\!A$} \\
\hline
CE & $59.87 \pm 0.76$ & $71.86 \pm 2.30$ \\
Feature KD & $55.48 \pm 2.85$ & $69.16 \pm 5.35$ \\
Vanilla KD & $57.91 \pm 4.93$ & $69.03 \pm 4.68$ \\
\hline
\textbf{UCMKD} & $\mathbf{72.76 \pm 5.14}$ & $\mathbf{74.35 \pm 1.27}$ \\
\hline
\end{tabular}
\end{table}
\textbf{Data Scarcity Scenario.}~To evaluate the robustness of our method under data-scarcity conditions, we conduct experiments using reduced training sets.~Specifically, we subsample the original training data at ratios of $0.3$, $0.4$, and $0.5$, while retaining the full test set to ensure a consistent evaluation benchmark. Tables~\ref{tab:ravdess_05}, \ref{tab:ravdess_04}, and \ref{tab:ravdess_03} report the performance on the {\bf RAVDESS} dataset of our method and two unpaired-setting baselines:~{\bf Cross-Entropy} and {\bf Feature Distillation}.~We also include {\bf Vanilla KD}~\citep{hinton2015distillingknowledgeneuralnetwork} under the paired setting using the same ResNet-18 backbone.~The results show that our method consistently achieves the strongest performance across all data-constrained scenarios.

\begin{table}[h]
\raggedright
\caption{Prediction accuracy on the {\bf RAVDESS} dataset using the ResNet-18 backbone with a training subset ratio of $0.4$, comparing our method {\bf UCMKD} against the {\bf Cross-Entropy}, {\bf Feature-Based Knowledge Distillation}, and {\bf Vanilla KD} baselines.}
\label{tab:ravdess_04}
\begin{tabular}{|lcc|}
\hline
\textbf{Method} & \textbf{$A\!\to\!V$} & \textbf{$V\!\to\!A$} \\
\hline
CE & $57.14 \pm 3.64$ & $69.93 \pm 3.75$ \\
Feature KD & $50.88 \pm 3.49$ & $71.96 \pm 0.84$ \\
Vanilla KD & $57.68 \pm 1.50$ & $71.63 \pm 0.91$ \\
\hline
\textbf{UCMKD} & $\mathbf{69.56 \pm 2.17}$ & $\mathbf{78.72 \pm 0.80}$ \\
\hline
\end{tabular}
\end{table}

\begin{table}[h]
\raggedright
\caption{Prediction accuracy on the {\bf RAVDESS} dataset using the ResNet-18 backbone with a training subset ratio of $0.3$, comparing our method against the {\bf Cross-Entropy}, {\bf Feature-Based Knowledge Distillation}, and {\bf Vanilla KD} baselines.}
\label{tab:ravdess_03}
\begin{tabular}{|lcc|}
\hline
\textbf{Method} & \textbf{$A\!\to\!V$} & \textbf{$V\!\to\!A$} \\
\hline
CE & $54.51 \pm 4.10$ & $68.96 \pm 2.20$ \\
Feature KD & $50.71 \pm 6.76$ & $63.97 \pm 2.20$ \\
Vanilla KD & $50.98 \pm 4.40$ & $68.46 \pm 0.60$ \\
\hline
\textbf{UCMKD} & $\mathbf{69.03 \pm 2.50}$ & $\mathbf{77.32 \pm 1.34}$ \\
\hline
\end{tabular}
\end{table}

\textbf{Parameter Sensitivity.}~We conduct a sensitivity analysis to evaluate the effect of the hyperparameters $\lambda_1$ and $\lambda_2$, which control the relative contributions of the Feature Alignment (\textbf{FA}) and Label Alignment (\textbf{LA}) losses, respectively.~Across these experiments, the inner-loop optimization steps are fixed to $n_1 = n_2 = 1$ (see Algorithm~\ref{algo:metaKD}).~Figure~\ref{fig:heat_map_lamda} presents the resulting performance heatmaps (audio-to-visual and visual-to-audio) on the {\bf RAVDESS} dataset~\citep{Liu2022DeepCross} according to the hyper-parameter grid with different choices for $\lambda_1$ and $\lambda_2$ within ${0.2, 0.4, 0.6, 0.8, 1.0}$.~The best performance for audio-to-visual ($A \to V$) distillation is achieved at $(\lambda_1, \lambda_2) = (0.2, 1.0)$, while visual-to-audio ($V \to A$) distillation performs best at $(\lambda_1, \lambda_2) = (1.0, 0.8)$.~Notably, even under less favorable hyperparameter configurations, our method consistently outperforms the baselines on $A \to V$ and remains competitive on $V \to A$.~These results demonstrate the robustness and stability of our {\bf UCMKD} across a broad range of loss-weight configurations.

\begin{figure}
    \raggedright
\includegraphics[width=1.0\linewidth]{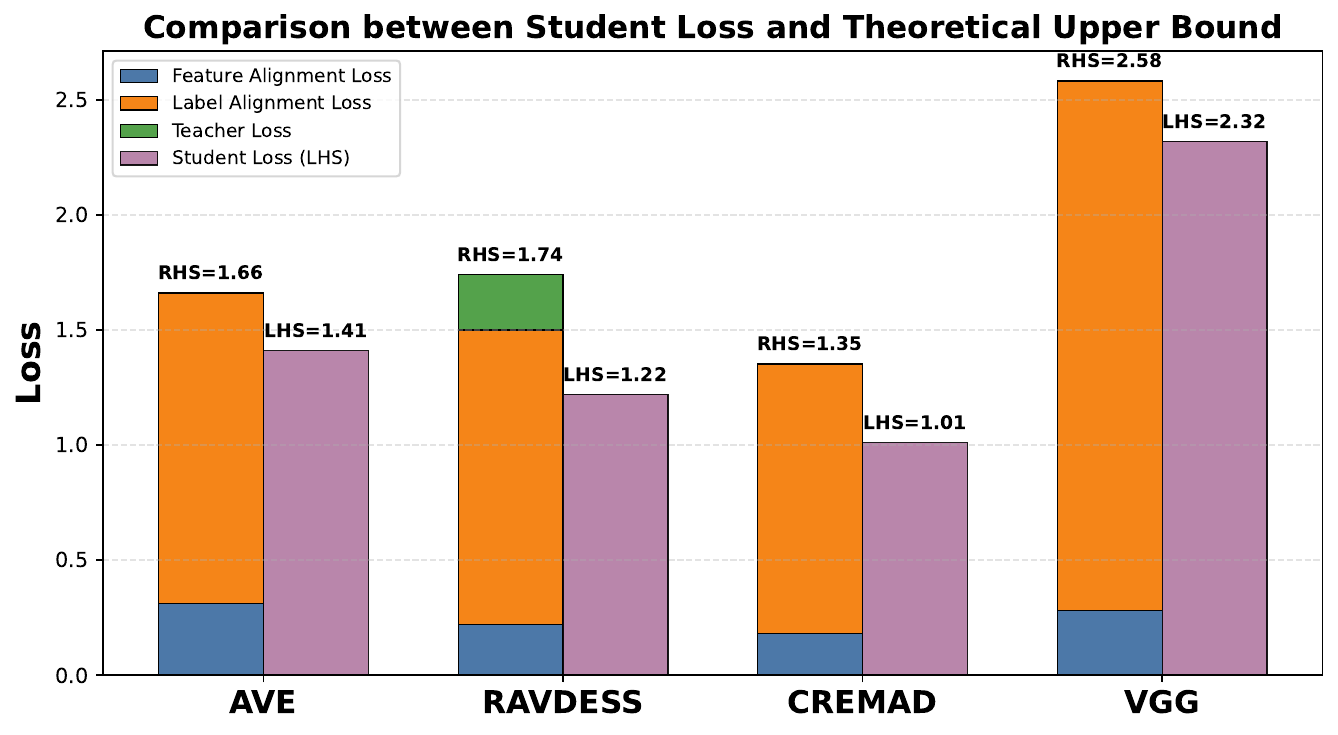}
    \caption{Informativeness of the theoretical bound across the {\bf AVE}, {\bf RAVDESS}, {\bf CREMA-D}, and {\bf VGGSound} datasets.~The proposed bound remains reasonably tight with an average gap of $24.5\%$.}
    \label{fig:tightness}
\end{figure}

\textbf{Informativeness of Theoretical Bounds.}~To empirically validate the informativeness (i.e., tightness) of the theoretical results developed in Section~\ref{sec:theory}, we evaluate the gap between the theoretical bounds and the observed empirical performance across multiple datasets.~Figure~\ref{fig:tightness} summarizes the resulting tightness analysis for both the infinite-data setting (Theorem~\ref{thm:2}) and the finite-data regime (Theorem~\ref{thm:3}).~Overall, the bounds remain reasonably tight across all datasets, with an average gap of $24.5\%$. Notably, on the large-scale {\bf VGGSound} dataset containing over 300K+ samples, the gap decreases to $11\%$, suggesting that the bounds become increasingly informative as data coverage grows.~This observation is consistent with the theoretical behavior characterized in Theorems~\ref{thm:2} and~\ref{thm:3}.

\textbf{Component-wise Contribution.}~Table~\ref{tab:component_contribution_ablation} presents an ablation study on the contributions of feature alignment (FA), label alignment (LA), and the bi-level optimization framework.~We observe that FA-only and LA-only variants each achieve competitive performance individually.~However, directly combining both alignment losses without the bi-level formulation leads to noticeable performance degradation, highlighting the instability of naive simultaneous optimization (see Section~\ref{sec:algo_design}).~In contrast, \textbf{UCMKD}, which integrates both alignment objectives within the proposed bi-level framework, consistently achieves the strongest performance across all settings.~These findings align with the theoretical insights of Theorem~\ref{thm:2} and demonstrate the importance of both alignment components and the bi-level optimization approach.

\begin{table}[t]
\raggedright
\caption{Component-wise ablation results on the {\bf AVE} and {\bf RAVDESS} datasets comparing FA-only (bi-level), LA-only (bi-level), both (w/o bi-level), and our {\bf UCMKD} framework.}
\label{tab:component_contribution_ablation}

\iffalse
\begin{tabular}{|lcccc|}
\hline
\multirow{2}{*}{Method} & \multicolumn{2}{c}{AVE} & \multicolumn{2}{c}{RAVDESS} \\
\cmidrule(lr){2-3} \cmidrule(lr){4-5}
 & A$\rightarrow$V & V$\rightarrow$A & A$\rightarrow$V & V$\rightarrow$A \\
\hline
FA-only (bilevel)   & 31.01 & 48.51 & 65.37 & 69.80 \\
LA-only (bilevel)   & 30.02 & 48.92 & 67.90 & 69.33 \\
Both (w/o bilevel)  & 28.11 & 48.26 & 66.47 & 69.53 \\
\textbf{UCMKD}      & \textbf{34.16} & \textbf{52.24} & \textbf{73.83} & \textbf{74.43} \\
\hline
\end{tabular}
\fi

\begin{tabular}{|l|c|c|c|c|}
\hline
\multirow{2}{*}{\bf Method} & \multicolumn{2}{c|}{\bf AVE} & \multicolumn{2}{c|}{\bf RAVDESS} \\
\cline{2-5}
 & A$\rightarrow$V & V$\rightarrow$A & A$\rightarrow$V & V$\rightarrow$A \\
\hline
FA-only (bilevel)   & 31.01 & 48.51 & 65.37 & 69.80 \\
LA-only (bilevel)   & 30.02 & 48.92 & 67.90 & 69.33 \\
Both (w/o bilevel)  & 28.11 & 48.26 & 66.47 & 69.53 \\
\textbf{UCMKD}      & \textbf{34.16} & \textbf{52.24} & \textbf{73.83} & \textbf{74.43} \\
\hline
\end{tabular}
\end{table}

\textbf{Alternative Distribution Distances.}~We note that using the $\ell_2$ cost for OT is standard practice in prior works~\citep{damodran2018DeepJ,tran2026rethinking}.~Moreover, our theoretical bound is agnostic to the specific choice of cost metric. To further assess this design choice, Table~\ref{tab:ot_cost_metrics} compares the performance of several alternative cost metrics.~Overall, all metrics achieve comparable results.~Among them, $\ell_2$ achieves marginally better results on 3 out of 4 tasks, making it a reasonable default choice in practice.

\begin{table}[t]
\raggedright
\caption{Comparison of different transport cost metrics, including $\ell_1$, $\ell_2$, and angular distance, on {\bf RAVDESS} and {\bf CREMA-D}.}
\label{tab:ot_cost_metrics}
\begin{tabular}{|lccc|}
\hline
Dataset & $\ell_1$ & $\ell_2$ & Angular \\
\hline
RAVDESS (A$\rightarrow$V) & \textbf{74.23} & 73.83 & 71.03 \\
RAVDESS (V$\rightarrow$A) & 73.65 & \textbf{74.44} & 73.03 \\
CREMA-D (A$\rightarrow$V) & 70.67 & \textbf{71.68} & 71.07 \\
CREMA-D (V$\rightarrow$A) & 63.52 & \textbf{66.59} & 62.41 \\
\hline
\end{tabular}
\end{table}

\textbf{Scalability with Larger Backbones.}~To further evaluate scalability under more realistic settings, we conduct additional experiments using ViT-based architectures, with ViT-B as the teacher and ViT-S as the student following the protocol in~\citep{Addepalli_2024_CVPR}.~These models are representative of modern large-scale vision architectures.~As shown in Table~\ref{tab:vit_scalability}, {\bf UCMKD} consistently achieves the best student performance across all datasets and transfer directions.~These results demonstrate that the proposed framework remains effective when transitioning from ResNet-based backbones to substantially larger ViT-based architectures.~Combined with the complexity analysis in Appendix~\ref{sec:complexity}, these findings provide further evidence of the scalability of our approach in practical large-scale settings.

\begin{table}[t]
\raggedright
\caption{Scalability evaluation using a ViT-based architecture with ViT-B (patch16-224, ~86M parameters) as the teacher and ViT-S (patch16-224, ~22M parameters) as the student.}
\label{tab:vit_scalability}
\iffalse
\begin{tabular}{|lcccc|}
\hline
\multirow{2}{*}{{\bf Method}} & \multicolumn{2}{c}{{\bf AVE}} & \multicolumn{2}{c}{{\bf RAVDESS}} \\
\cmidrule(lr){2-3} \cmidrule(lr){4-5}
 & A$\rightarrow$V & V$\rightarrow$A & A$\rightarrow$V & V$\rightarrow$A \\
\hline
Teacher          & 75.87 & 70.15 & 90.41 & 89.11 \\
CE               & 51.19 & 53.73 & 65.63 & 66.13 \\
Feature KD  & 50.96 & 56.22 & 69.83 & 67.73 \\
\textbf{UCMKD}   & \textbf{56.97} & \textbf{58.21} & \textbf{80.32} & \textbf{72.43} \\
\hline
\end{tabular}
\fi
\begin{tabular}{|l|c|c|c|c|}
\hline
\multirow{2}{*}{\bf Method} & \multicolumn{2}{c|}{\bf AVE} & \multicolumn{2}{c|}{\bf RAVDESS} \\
\cline{2-5}
 & A$\rightarrow$V & V$\rightarrow$A & A$\rightarrow$V & V$\rightarrow$A \\
\hline
Teacher          & 75.87 & 70.15 & 90.41 & 89.11 \\
CE               & 51.19 & 53.73 & 65.63 & 66.13 \\
Feature KD       & 50.96 & 56.22 & 69.83 & 67.73 \\
\textbf{UCMKD}   & \textbf{56.97} & \textbf{58.21} & \textbf{80.32} & \textbf{72.43} \\
\hline
\end{tabular}
\end{table}

\section{Related Works}
\label{sec:related}
In this section, we provide the literature review of
the most relevant works on unimodal KD (Section~\ref{sec:related_work_uni_kd}) and cross-modal KD (Section~\ref{sec:related_work_crossmoda_kd}). The detailed formulations of both mentioned settings are provided in Appendix ~\ref{sec:addtional_related_works}.
\subsection{In-Modal Knowledge Distillation}
\label{sec:related_work_uni_kd}
In-Modal Knowledge Distillation (KD) transfers the knowledge of a pretrained teacher model to a student model by minimizing discrepancies between their output predictions or intermediate representations.~The seminal work of \citep{hinton2015distillingknowledgeneuralnetwork} formulates this as minimizing the Kullback–Leibler (KL) divergence between teacher and student soft predictions, enabling the compression of large models while largely preserving predictive performance.~Follow-up work has explored a broad range of alternative mechanisms for effective knowledge transfer.~For example, CRD~\citep{Tian2020Contrastive} introduces contrastive objectives for representation-level distillation, while SCKD~\citep{Zhu2021SCKD} adaptively adjusts the distillation process according to gradient similarity between teacher and student models.~DKD~\citep{Zhao2022DKD} decomposes KD into target-class and non-target-class distillation to improve flexibility, and Review~\citep{Chen2021Review} leverages multi-level teacher representations through a feature-review mechanism.~DIST~\citep{huang2022knowledgedistillationstrongerteacher} further develops correlation-based objectives to capture inter-class and intra-class structural relations, while L2D~\citep{Yang2023MLKD} extends such relation-based distillation to multi-label classification.~SHAKE~\citep{li2022shadow} bridges offline and online KD through auxiliary shadow heads, \citep{lv2024wasserstein} replaces the KL divergence with Wasserstein-distance-based objectives, and RLD~\citep{sun2025knowledgedistillationrefinedlogits} dynamically refines teacher logits using label information to suppress misleading supervision from incorrect teacher predictions.

Despite their strong empirical performance, these methods predominantly assume that teacher and student models operate on identical training data, thereby overlooking the challenges of Cross-Modal Knowledge Distillation (CMKD), particularly under unpaired settings where explicit sample-level correspondence is unavailable.%\vspace{-2mm}

\subsection{Cross-Modal Knowledge Distillation}
\label{sec:related_work_crossmoda_kd}
Cross-Modal Knowledge Distillation (CMKD) extends traditional KD to settings where teacher and student models operate on different data modalities.~Early work by \citet{gupta2015crossmodaldistillationsupervision} transfers supervision from a labeled modality to an unlabeled paired modality, while \citet{Roheda2018Cross} employs GANs to transfer knowledge across missing and available modalities.~\citet{Xue2021MKEx} adapts multimodal networks to unlabeled modalities through pseudo-label sampling from unimodal teachers, and \citet{Lee2023Decomp} develops decomposed cross-modal distillation for RGB-based detection using optical-flow supervision.

Other work explores cross-modal transfer for dense indoor prediction~\citep{yun2023dense2d3dindoorprediction} and theoretical understanding of modality interactions through modality Venn diagrams and modality-focusing hypotheses~\citep{xue2023modalityfocusinghypothesisunderstanding}.~More recent approaches aim to improve cross-modal transfer under increasingly complex settings. $\text{C}2\text{KD}$~\citep{Huo2024C2KD} introduces selective bidirectional distillation to bridge modality gaps, while \citep{sarkar2023xkdcrossmodalknowledgedistillation} develops a self-supervised framework for cross-modal distillation from unlabeled videos. COSMOS~\citep{kim2025cosmoscrossmodalityselfdistillationvision} further incorporates text-cropping and cross-attention mechanisms for vision-language models, and XKD~\citep{bendidi2025a} explores weakly paired microscopy and transcriptomics data.

Despite these advances, existing CMKD methods still rely heavily on paired or weakly paired multi-modal data, which are often costly or unavailable in realistic settings.~Addressing this has been the focus of this paper.%\vspace{-2mm}

%~In this work, we move beyond explicit sample-level correspondence by establishing a distributional relationship between teacher and student models.~Building on this formulation, we aim to develop a principled framework with theoretical guarantees that enables cross-modal distillation through distribution-level alignment rather than paired-data supervision.

\section{Limitations and Future Works}
\label{sec:limitation}
While our framework provides a principled formulation and effective algorithm for unpaired CMKD, several directions remain open.~First, although our theoretical analysis permits arbitrary transport costs $\delta(\cdot,\cdot)$, we instantiate the framework using the Euclidean metric for simplicity.~Learning the transport geometry via adaptive metric learning may further tighten the alignment objective.~Second, the proposed bi-level optimization introduces additional computational overhead due to second-order updates or their approximations. Improving optimization efficiency through implicit differentiation or reduced unrolling would further enhance scalability. Third, our results on the large-scale {\bf VGGSound} benchmark suggest that the framework remains effective beyond small curated datasets, motivating future exploration in settings with foundation models where cross-modal transfer must operate without explicit sample-level pairing.~Finally, since the proposed framework relies on distribution-level alignment rather than paired supervision, extending it to cross-modal generative modeling under unpaired settings is another promising direction for future follow-up.%\vspace{-2mm}

%While our framework provides a principled view of unpaired CMKD and a practical algorithm, several directions remain open. First, our theoretical bound allows any admissible cost metric $\delta(\cdot, \cdot)$, but in practice we instantiate it with a Euclidean cost. This choice is convenient and does not affect the validity of the bound, yet it may be suboptimal for tightening it. A natural extension is to learn the transport geometry by parameterizing the cost with a metric-learning module and optimizing it jointly with the student. Second, the proposed bi-level updates involve second-order information (or approximations), which increases training cost. Improving efficiency, implicit differentiation, or reduced unrolling would make the method more scalable. Third, for large-scale and foundation model settings, our results on VGGSound, a large-scale benchmark, suggest that the approach remains effective beyond small curated datasets and is compatible with regimes where data are abundant and heterogeneous. This points to a natural next step: applying unpaired CMKD as an alignment and transfer mechanism for foundation models, where a large multimodal model can be distilled into lightweight students or adapted across modalities without requiring instance-level pairing. Finally, because our approach is built on distribution matching and does not rely on paired samples, it is well-suited to settings beyond classification. In particular, extending the alignment objectives to cross-modal generation (e.g., diffusion-based models) under unpaired data is an exciting direction.
\section{Conclusion}%\vspace{-1mm}
This paper studies cross-modal knowledge distillation under the challenging unpaired setting and proposes \textbf{UCMKD}, a principled framework built upon two key components: {\bf Feature Alignment} and {\bf Label Alignment}.~We establish both infinite-sample and finite-sample generalization bounds for the student model and develop a practical meta-learning-style optimization framework to realize these objectives. Extensive experiments on {\bf AVE}, {\bf RAVDESS}, {\bf CREMA-D}, and {\bf VGGSound} demonstrate consistent improvements in unpaired settings while remaining competitive with SOTA methods when paired data are available.~The proposed formulation also naturally extends beyond prediction task to general distribution-matching losses, opening promising directions for large-scale multi-modal transfer and unsupervised cross-modal generative modeling.\vspace{-2mm}

%To conclude, we study knowledge distillation in the challenging unpaired setting and propose \textbf{UCMKD}, a principled framework built on two core components: Feature Alignment and Label Alignment. We derive both infinite-sample and finite-sample generalization bounds for the student and introduce a practical meta-learning style optimization procedure to realize these objectives. Experiments on AVE, RAVDESS, CREMA-D, and VGGSound demonstrate consistent gains in unpaired scenarios while remaining competitive with state-of-the-art methods when paired data are available. Beyond classification, our formulation naturally supports broader distribution-matching objectives, suggesting extensions to large-scale multimodal transfer and generative settings under missing or unpaired supervision.

\section*{Acknowledgement}\vspace{-1mm}
This work utilized GPU compute resources at SDSC and ACES through allocation CIS230391 from the Advanced Cyberinfrastructure Coordination Ecosystem:~Services and Support (ACCESS) program~\cite{ACCESS-resource}, which is supported by U.S. National Science Foundation grants $\#$2138259, $\#$2138286, $\#$2138307, $\#$2137603, and $\#$2138296.~Trong Nghia Hoang is supported by National Science Foundation CAREER Award IIS-2544071.~The authors also acknowledge the compute support from Modal.

\newpage

\section*{Impact Statement}
Our work provides a framework for cross-modal knowledge distillation without requiring paired data, enabling multimodal transfer in settings where synchronized datasets are unavailable.~However, distribution-level alignment might propagate biases or spurious correlations inherited from the teacher model.~Careful inspection of the teacher model is important when such systems are deployed in practice.

%Our work offers a principled framework for cross-modal knowledge distillation without paired data, helping researchers and practitioners leverage multimodal intelligence even in data-scarce environments where collecting synchronized datasets is infeasible. However, relying on distributional alignment rather than instance-level supervision carries the risk of propagating latent biases from the teacher model or learning spurious correlations. We encourage the community to carefully validate teacher quality and distribution compatibility when deploying these methods, ensuring that the gain in data efficiency does not compromise model fairness or reliability.

%\newpage

\bibliography{example_paper}
\bibliographystyle{icml2026}

%%%%%%%%%%%%%%%%%%%%%%%%%%%%%%%%%%%%%%%%%%%%%%%%%%%%%%%%%%%%%%%%%%%%%%%%%%%%%%%
%%%%%%%%%%%%%%%%%%%%%%%%%%%%%%%%%%%%%%%%%%%%%%%%%%%%%%%%%%%%%%%%%%%%%%%%%%%%%%%
% APPENDIX
%%%%%%%%%%%%%%%%%%%%%%%%%%%%%%%%%%%%%%%%%%%%%%%%%%%%%%%%%%%%%%%%%%%%%%%%%%%%%%%
%%%%%%%%%%%%%%%%%%%%%%%%%%%%%%%%%%%%%%%%%%%%%%%%%%%%%%%%%%%%%%%%%%%%%%%%%%%%%%%
\newpage
\appendix
\onecolumn
\section{Proof for the infinity data points case.}
\label{app:A}
We have the generalized error for the teacher model and the student model as: 
\begin{equation}
    \text{err}_{T} =  \mathbb{E}_{D^T(\boldsymbol{x}^T, y)} \Big[- \log p_T(y \mid \boldsymbol{z} = \theta(\boldsymbol{x}^T)) \Big]
\end{equation}
\begin{equation}
    \text{err}_{S}  = \mathbb{E}_{D^S(\boldsymbol{x}^S, y)} \Big[- \log p_S(y \mid \boldsymbol{z} = \phi(\boldsymbol{x}^S)) \Big] 
\end{equation}
We denote the joint distributions: 
\begin{equation}
    D^T(\theta(\boldsymbol{x}^T), y) := D^{T}(\boldsymbol{z}, y) = D^T( \boldsymbol{z}) D^T(y \mid \boldsymbol{z})
\end{equation}
\begin{equation}
    D^S(\phi(\boldsymbol{x}^S), y) := D^S(\boldsymbol{z}, y) = D^S(\boldsymbol{z}) D^S(y \mid \boldsymbol{z}) 
\end{equation}
Then, we can rewrite the generalized error of the teacher and the student model as: 
\begin{equation}
    \text{err}_T = \mathbb{E}_{D^{T}(\boldsymbol{z}, y)} \Big[-\log (p_T(y \mid \boldsymbol{z})) \Big] = \mathbb{E}_{D^T(\boldsymbol{z})} \mathbb{E}_{D^T(y \mid \boldsymbol{z})} \Big[-\log (p_T( y \mid \boldsymbol{z})) \Big]
\end{equation}
\begin{equation}
    \text{err}_S = \mathbb{E}_{D^{S}(\boldsymbol{z}, y)} \Big[-\log (p_S(y \mid \boldsymbol{z})) \Big] = \mathbb{E}_{D^S(\boldsymbol{z})} \mathbb{E}_{D^S(y \mid \boldsymbol{z})} \Big[-\log (p_S(y \mid \boldsymbol{z})) \Big]
\end{equation}
We have: 
\begin{equation}
    \text{err}_{S} - \text{err}_{T} = \textbf{A} + \textbf{B}
\end{equation}
where: 
\begin{align}
    \mathbf{A}:&=&  &\text{err}_S - \mathbb{E}_{D^S(\boldsymbol{z})} \mathbb{E}_{D^T(y \mid \boldsymbol{z})} \Big[ - \log(p_T(y  \mid \boldsymbol{z}))\Big] \\ 
    \mathbf{B}:&=& &\mathbb{E}_{D^S(\boldsymbol{z})} \mathbb{E}_{D^T(y \mid \boldsymbol{z})} \Big[ - \log(p_T(y \mid \boldsymbol{z}))\Big] - \text{err}_T
\end{align}
\noindent {\bf \underline{1.~Bounding $\mathbf{A}$.}} We have: 
\begin{align}
    \mathbf{A} &=&  &\mathbb{E}_{D^S(\boldsymbol{z})} \mathbb{E}_{\mathcal{D}^S(y \mid \boldsymbol{z})} \Big[-\log (p_S(y \mid \boldsymbol{z})) \Big] - \mathbb{E}_{D^S(\boldsymbol{z})} \mathbb{E}_{D^T(y \mid \boldsymbol{z})} \Big[ - \log(p_T(y \mid \boldsymbol{z}))\Big] \\
     &=& &\mathbb{E}_{D^S(\boldsymbol{z})} \Big[ \mathbb{E}_{D^S(y \mid \boldsymbol{z})} \Big[-\log (p_S(y \mid \boldsymbol{z})) \Big] - \mathbb{E}_{D^T(y \mid \boldsymbol{z})} \Big[ - \log(p_T(y \mid \boldsymbol{z}))\Big]\Big] \\
     &=& &\mathbb{E}_{D^S(\boldsymbol{z})} \Big[-\sum_{y \in \mathcal{Y}} \Big(D^S(y \mid \boldsymbol{z}) \log (p_S(y \mid \boldsymbol{z})) - D^{T}(y \mid \boldsymbol{z}) \log(p_T(y \mid \boldsymbol{z}))\Big)\Big]  \\
     &=& &\mathbb{E}_{D^{S}(\boldsymbol{z})} \mathbb{E}_{D^S(y \mid \boldsymbol{z})} \Big[ -\log(p_S(y \mid \boldsymbol{z})) + \frac{D^{T}(y \mid \boldsymbol{z})}{D^{S}(y \mid \boldsymbol{z})} \log(p_T(y \mid \boldsymbol{z}))\Big] \label{term_A}
\end{align}
With a mild assumption $D^{S}(y \mid \boldsymbol{z})>0$, we denote the label transport kernel $\kappa(y,\boldsymbol{z}) \triangleq \frac{D^{T}(y \mid \boldsymbol{z})}{D^{S}(y \mid \boldsymbol{z})} $. Thus, term $\mathbf{A}$ then can be expressed as: 
\begin{equation}
    \mathbf{A} = \mathbb{E}_{D^S(\boldsymbol{z},y)} \Big[- \log\Big(\frac{p_S(y \mid \boldsymbol{z})}{p_T(y \mid \boldsymbol{z})^{\kappa(y, \boldsymbol{z})}} \Big) \Big]
\end{equation}

\noindent {\bf \underline{2.~Bounding $\mathbf{B}$.}} We have: 
\begin{align}
    \mathbf{B} &=&  &\mathbb{E}_{D^S(\boldsymbol{z})} \mathbb{E}_{D^T(y \mid \boldsymbol{z})} \Big[ - \log(p_T(y \mid \boldsymbol{z}))\Big] - \mathbb{E}_{D^T(\boldsymbol{z})} \mathbb{E}_{D^T(y \mid \boldsymbol{z})} \Big[-\log (p_T(y \mid \boldsymbol{z})) \Big] \\ 
     &=&  &\mathbb{E}_{D^S(\boldsymbol{z})} \Big[\ell_\tau(\boldsymbol{z}) \Big] -  \mathbb{E}_{D^T(\boldsymbol{z})} \Big[\ell_\tau(\boldsymbol{z}) \Big]
\end{align}
where $\ell_\tau(\boldsymbol{z}) \triangleq \mathbb{E}_{D^T(y \mid \boldsymbol{z})} \Big[ - \log(p_T(y \mid \boldsymbol{z}))\Big]$ is the cross-entropy of the teacher prediction as Definition~\ref{def:FA}. For any cost metric $\delta \in \Delta$  such that $|\ell_\tau(\boldsymbol{z}_1) - \ell_\tau(\boldsymbol{z}_2)| \leq \tau_{\delta} \cdot\delta(\boldsymbol{z}_1, \boldsymbol{z}_2)$, the Kantorovich-Rubinstein duality ascertains that:
\begin{equation}
    \mathbf{B} \leq \tau_{\delta} \mathbf{W}_{\delta}\Big(D^{S}(\boldsymbol{z}), D^{T}(\boldsymbol{z})\Big) \label{term_B}
\end{equation}
where $\mathbf{W}_{\delta}$ denotes the Wasserstein $-1$ distance with the cost metric $\delta$. Combine Eq.~\eqref{term_A} and Eq.~\eqref{term_B} we have: 
\begin{equation}
    \text{err}_S \leq \text{err}_T + \tau_{\delta} \mathbf{W}_{\delta}\Big(D^{S}(\boldsymbol{z}), D^{T}(\boldsymbol{z})\Big) + \mathbb{E}_{D^S(\boldsymbol{z},y)} \Big[- \log\Big(\frac{p_S(y \mid \boldsymbol{z})}{p_T(y \mid \boldsymbol{z})^{\kappa(y,\boldsymbol{z})}} \Big) \Big]
\end{equation}
Using Definition~\ref{def:FA} and Definition~\ref{def:LA}, finally, we complete our proof: 
\begin{align}
    \text{err}_{S} \leq \text{err}_{T} + \textbf{FA}(\theta, \phi) + \textbf{LA}(p_S, p_T) 
\end{align}

\section{Proof for the finite data points case.}
\label{app:B}
\noindent {\bf \underline{1.~Rademacher Bounds.}} 

We start with the Rademacher bound \citep{koltchinskii2004rademacherprocessesboundingrisk}, which is stated as follows.

\textbf{Rademacher Bounds}. Let $\mathcal{F}$ is the family of functions mapping from $Z$ to $[0,1]$. Then for any $0< \delta < 1$, with probability at least $1- \delta$ over sample $S= \{z_1, \cdot \cdot \cdot, z_n \}$, the following holds for all $f \in \mathcal{F}$: 
\begin{eqnarray}
    \mathbb{E}[f] &\leq& \frac{1}{n} \sum_{i=1}^{n} f(z_i) + 2 \mathcal{R}_{n}(\mathcal{F}) +  \sqrt{\frac{\log(1/ \delta)}{2n}}  \\ 
    \mathbb{E}[f] &\leq& \frac{1}{n} \sum_{i=1}^{n} f(z_i) + 2 \hat{\mathcal{R}}_{S}(\mathcal{F}) +  3\sqrt{\frac{\log(2/ \delta)}{2n}}
\end{eqnarray}
Where $\mathcal{R}_n(\mathcal{F})$ and $\hat{\mathcal{R}}_{S}(\mathcal{F})$ are the Rademacher complexity and the empirical Rademacher complexity.

\noindent {\bf \underline{2.~Bounding Wasserstein distance.}}  

Feature Alignment (\text{FA}) is formulated as \textbf{Wasserstein Distance} with the momentum $ p= 1$, cost metric $\delta$, and high dimension $d>1$. For clear notation, we introduce two true probability distributions $\nu$ and $\mu$ with their empirical distributions $\hat{\nu}_n$ and $\hat{\mu}_m$ which provided by $n$ and $m$ data points, respectively. Using the triangle inequality, the Wasserstein distance term can be expressed as: 
\begin{equation}
     \mathbf{W}(\nu,\mu) \leq \mathbf{W}(\hat{\nu}_n, \hat{\mu}_m) + \mathbf{W}(\nu, \hat{\nu}_n) + \mathbf{W}(\mu, \hat{\mu}_m)
\end{equation}
Where $\mathbf{W}(\hat{\nu}_n, \hat{\mu}_m)$ can be seen as the empirical estimation of $\mathbf{W}(\nu, \mu)$, which can be directly computed using standard numerical methods. Next, we will explore $\mathbf{W}(\nu, \hat{\nu}_n)$ term as well as $\mathbf{W}(\mu, \hat{\mu}_m)$ term in context \textbf{Wasserstein$-1$ Distance}.

For all $n>0$ and finite momentum $1 \leq p < \infty$, \citep{weed2017sharpasymptoticfinitesamplerates} stated that: 
\begin{equation}
    \mathbb{P}\Big(\mathbf{W}(\nu, \hat{\nu}_n) - \mathbb{E} \big[\mathbf{W}(\nu, \hat{\nu}_n)\big] \geq t  \Big) \leq \text{exp}(-2nt^2)
\end{equation}
Thus, with the probability at least $1- \text{exp}(-2nt^2)$, we have: 
\begin{equation}
    \mathbf{W}(\nu, \hat{\nu}_n) \leq \mathbb{E} \big[\mathbf{W}(\nu, \hat{\nu}_n)\big] + t
\end{equation}
We then define $d^*_p(\nu)$ as the upper Wasserstein dimensions \cite{weed2017sharpasymptoticfinitesamplerates}. Using Theorem 1 in \cite{weed2017sharpasymptoticfinitesamplerates}, given the finite momentum $1 \leq p< \infty$ and $s_1> d_{p}^{*}(\nu)$ is the upper Wasserstein dimension, exist a constant $C_1 > 0$, such that:
\begin{equation}
    \mathbb{E} \big[\mathbf{W}(\nu, \hat{\nu}_n)\big] \leq C_{1} n^{-1/s_1}
\end{equation}
Thus, denote $\delta \triangleq \text{exp}(-2nt^2)$ and $0<\delta<1$, with the probability at least $1-\delta$, we have:  
\begin{equation}
     \mathbf{W}(\nu, \hat{\nu}_n) \leq C_1 n^{-1/s_1} + \sqrt{\frac{\log(2/\delta)}{2n}}
\end{equation}
Deriving the same steps, with the probability at least $1-\delta$, we have: 
\begin{equation}
     \mathbf{W}(\mu, \hat{\mu}_m) \leq C_2 m^{-1/s_2} + \sqrt{\frac{\log(2/\delta)}{2m}}
\end{equation}
Finally, with the probability at least $1- 2\delta$, the Wasserstein distance can be bounded by: 
\begin{align}
    \label{WS_bound}
    \textbf{W}(\nu, \mu) \leq \mathbf{W}(\hat{\nu}_n, \hat{\mu}_m) + C_1 n^{-1/s_1} +C_2 m^{-1/s_2} + \sqrt{\frac{\log(2/\delta)}{2n}} + 
      \sqrt{\frac{\log(2/\delta)}{2m}} 
\end{align}
where $C_1, C_2, s_1,s_2$ are positive constants, $s_1,s_2$ are larger than the upper Wasserstein dimensions of $\nu$ and $\mu$, respectively.    

\noindent {\bf \underline{3.~Bounding Label Alignment.}}    

Denoting $f(\boldsymbol{z},y) \triangleq - \log\Big(\frac{p_S(y \mid \boldsymbol{z})}{p_T(y \mid \boldsymbol{z})^{\kappa(y, \boldsymbol{z})}} \Big)$, we then express Label Alignment (\textbf{LA}) as: 
\begin{equation}
  \mathbf{LA} \triangleq  \mathbb{E}_{D^S(\boldsymbol{z},y)} \Big[- \log\Big(\frac{p_S(y \mid \boldsymbol{z})}{p_T(y \mid \boldsymbol{z})^{\kappa(y,\boldsymbol{z})}} \Big) \Big]  =  \mathbb{E}_{D^S(\boldsymbol{z},y)} \Big[f(\boldsymbol{z},y) \Big]
\end{equation}
With the mild assumption that the class function $f \in \mathcal{F}$ is upper-bounded by a constraint $C_3>0$, we can scale the function $f$ to $[0,1]$ by dividing by $C_3$ and denote the new class function as $\mathcal{F}/C_3$. Using Rademacher bound \citep{koltchinskii2004rademacherprocessesboundingrisk}, given $0<\delta<1$, with the the probability at least $1-\delta$ over $m$ provided sample, we have:
\begin{equation}
    \frac{\mathbb{E}[f] }{C_3}\leq \frac{\hat{E}[f]}{C_3} + 2 \mathcal{R}_{m}(\mathcal{F}/C_3)  +  \sqrt{\frac{\log(1/\delta)}{2m}} \label{LA_bound}
\end{equation}
where $\hat{E}[f] \triangleq \frac{1}{m}\sum_{i=1}^{m} f(\boldsymbol{z}_i, y_i)$ and $\mathcal{R}_{m}(\mathcal{F}/C_3)$ is Rademacher complexity. By using the property $\alpha \cdot \mathcal{R}(\mathcal{G}) = \mathcal{R}( \alpha \cdot \mathcal{G})$, we have: 

\begin{equation}
     \mathbb{E}[f]\leq \hat{E}[f] + 2 \mathcal{R}_{m}(\mathcal{F})  +  C_3\sqrt{\frac{\log(1/\delta)}{2m}}
\end{equation}

Let $\Pi_{\mathcal{F}}: \mathbb{N} \xrightarrow{} \mathbb{N}$ be the growth function. Applying Massart's lemma to $\mathcal{R}_{m}(\mathcal{F})$ \citep{mohri2012foundations}, we have: 
\begin{equation}
    \mathcal{R}_{m}(\mathcal{F}) \leq C_3\sqrt{\frac{2 \log \Pi_{\mathcal{F}}(m)}{m}}
\end{equation}

Let $d \triangleq \text{VCdim}(\mathcal{F})$ be the VC dimension of the hypothesis class function $\mathcal{F}$.  For all $m \in \mathbb{N}$, using Sauer’s lemma \cite{mohri2012foundations} we have: 
\begin{equation}
    \Pi_{\mathcal{F}}(m) \leq \sum_{i=0}^{d} \binom{m}{i}
\end{equation}
Then, for all $d\leq n$, we have:
\begin{equation}
    \Pi_{\mathcal{F}}(m) \leq \Big( \frac{em}{d}\Big)^{d}
\end{equation}
Finally, given $0<\delta<1$, the function class $f$ is upper-bounded by a constraint $C_3>0$, the V-C dimension $d$, with the probability at least $1-\delta$, we have: 
\begin{equation}
    \mathbb{E}[f] \leq \hat{E}[f] + 2C_3 \sqrt{\frac{2d \log(m/d)}{m}} + C_3 \sqrt{\frac{\log(1/\delta)}{2m}}
\end{equation}

\noindent {\bf \underline{4.~Bounding the generalized student error on Offline CMKD.}} 

In the Offline CMKD setting, the teacher error is fixed due to the fixed teacher backbone during the distillation process. We can treat the teacher's error as the fixed overhead, then combining E.q~\eqref{WS_bound}, and E.q~\eqref{LA_bound}, given $0 \leq \delta \leq 1/3$, the teacher and the student empirical distribution $D_{n_{T}}^{T}(\boldsymbol{z})$ and $D_{n_S}^{S}(\boldsymbol{z})$ provided by $n_T$ and $n_S$ data points respectively. Let the Monte Carlo estimation of the Label Alignment (LA) be $\mathbf{LA}_{e}(p_S,p_T)$, $s_1$ and $s_2$ are larger than the upper-bound Wasserstein dimensions \citep{weed2017sharpasymptoticfinitesamplerates} of the student and teacher representation distribution, respectively, with probability at least $1- 3\delta$, we have: 
\begin{align}
    \text{err}_{S} &\leq& &\text{err}_{T}   + \mathbf{LA}_{e}(p_S,p_T) + 2C_3 \sqrt{\frac{2d\log(n_S/d) }{n_S}} + C_3 \sqrt{\frac{\log(1/\delta)}{2n_S}}\\ \nonumber
    &+& &\tau_\delta \Big( \mathbf{W}\big(D_{n_T}^{T}(\boldsymbol{z}), D_{n_S}^{S}(\boldsymbol{z})\big) + C_1 n_S^{-1/s_1}  + C_2 n_T^{-1/s_2}  + \sqrt{\frac{\log(2/\delta)}{2n_S}} + \sqrt{\frac{\log(2/\delta)}{2n_T}} \Big)
\end{align}
We completed our proof.
\section{Knowledge Distillation additional formulation}
\label{sec:addtional_related_works}
In this section, we provide additional detailed formulation about Knowledge Distillation in both common settings: unimodal KD (Section \ref{sec:unimodal_KD_additional}) and cross-modal KD (Section \ref{sec:cross-modal_KD_addtional}). 
\subsection{In-Modal Knowledge Distillation}
\label{sec:unimodal_KD_additional}
Formally, we consider the $K$ -classes classification problem where both the teacher model and the student model receive the same input modality $X$ and produce the logit prediction over $K$ classes. Let $h_{\theta}(X)$ and $h_{\phi}(X)$ be the pre-softmax logit of the teacher model and the student model, respectively. Given a temperature $T$, we have the softened predictions as 
\begin{eqnarray}
     f_{\theta}(X;T) &=& \text{softmax}(h_{\theta}(X)/ T)  \\ \nonumber
     f_{\phi}(X;T) &=& \text{softmax}(h_{\phi}(X)/ T)
\end{eqnarray}
The student model is trained to minimize the weighted combination of cross entropy loss with respect to the ground-truth labels $Y$ and the distillation loss: 
\begin{eqnarray}
    \mathcal{L} = \lambda \text{CE}(f_{\phi}(X), Y) + (1 -\lambda) T^{2} \cdot\text{KL}\big( f_{\theta}(X;T)\mid \mid  f_{\phi}(X;T)\big) \label{eq:unimodal_KD}
\end{eqnarray}
where KL is the Kullback–Leibler divergence and $\lambda \in [0,1]$. The objective of the combined loss function is to encourage the small, simple student model to mimic the behavior of large, complex teacher model, thus enabling the compression of the large models while preserving performance \citep{hinton2015distillingknowledgeneuralnetwork}.  
\subsection{Cross-Modal Knowledge Distillation}
\label{sec:cross-modal_KD_addtional}
Cross-modal Knowledge Distillation generalizes the unimodal framework to heterogeneous modalities, allowing a teacher with
access to a stronger modality to guide a student with a weaker one. We consider two modalities, denoted by $X_1$ and $X_2$ processed by the teacher and student models, respectively, and a single label $Y$ for both. In this setting, $X_1$ and $X_2$ come from the same instance and have the same label, thus called \textbf{paired data} setting. The combined objective function extends from Eq.~\eqref{eq:unimodal_KD} as \citep{Liu2022DeepCross}:   
\begin{eqnarray}
    \mathcal{L} = \lambda \text{CE}(f_{\phi}(X_2), Y) + (1 -\lambda) T^{2} \cdot\text{KL}\big( f_{\theta}(X_1;T)\mid \mid  f_{\phi}(X_2;T)\big) 
\end{eqnarray}
\section{Practical Estimation of the Label Transport Kernel}
\label{sec:estimate_kernel}
The theoretical analysis introduces a label-transport kernel $\kappa(y, \boldsymbol{z}) \triangleq D^T(y \mid \boldsymbol{z})/D^S(y \mid \boldsymbol{z})$ to modulate label alignment in the absence of paired samples. In practical implementation, $\kappa$ is only required under the student conditional $y \sim D^S(\cdot \mid \boldsymbol{z})$. For supervised classification, the empirical conditional induced by the labeled student dataset is a Dirac distribution $\hat{D}^S(y \mid \boldsymbol{z}_i) = \delta(y = y_i)$ when given data point $(\boldsymbol{x}_i, y_i)$ under the feature map $\boldsymbol{z}_i = \phi(\boldsymbol{x}_i)$. Consequently, the kernel is evaluated only at the realized label $y_i$, such that 
\begin{equation}
    \hat{\kappa}_i = \kappa(y_i,\boldsymbol{z}_i) = \frac{\hat{D}^{T}(y_i \mid \boldsymbol{z}_i)}{ \hat{D}^S(y_i \mid \boldsymbol{z}_i)} = \hat{D}^{T}(y_i \mid \boldsymbol{z}_i)
\end{equation}
Given a good pre-trained teacher model, we further adopt a plug-in estimator $\hat{D}^T(y_i \mid \boldsymbol{z}_i) = p_T(y_i \mid \boldsymbol{z}_i)$ (pseudo label sampling \citep{nguyen2020leepnewmeasureevaluate}). Therefore, $\hat{\kappa}_i = p_T(y_i \mid \boldsymbol{z}_i)$ acts as a teacher–student label-compatibility score: when the teacher assigns low probability to the student’s ground-truth label, distillation is downweighted to mitigate negative transfer.
\section{Complexity Analysis}
\label{sec:complexity}
As discussed in Section~\ref{sec:limitation}, the bilevel optimization in UCMKD introduces additional training cost. However, this overhead mainly comes from a constant-factor increase in the number of student forward/backward passes. It does not introduce a new dependence on the number of training samples beyond the cost already required for computing the alignment objective. Therefore, the scaling behavior with respect to dataset size remains comparable to feature-based knowledge distillation baselines.

\paragraph{Empirical training cost.}
We first report the per-epoch training time of UCMKD and Feature-based KD in Table~\ref{tab:training_time}. Across datasets and transfer directions, UCMKD increases the per-epoch training time by a moderate constant factor, ranging from $1.20\times$ to $2.91\times$. This overhead is acceptable given the performance gains over existing baselines reported in Tables~\ref{tab:paired_results} and~\ref{tab:unpaired_results}.

\begin{table}[t]
\centering
\caption{Per-epoch training time comparison between UCMKD and Feature-based Knowledge Distillation. The relative overhead is computed as the ratio between the training time of UCMKD and Feature Distill.}
\label{tab:training_time}
\resizebox{0.8\linewidth}{!}{
\begin{tabular}{lcccccccc}
\toprule
\multirow{2}{*}{Method}
& \multicolumn{2}{c}{AVE}
& \multicolumn{2}{c}{CREMA-D}
& \multicolumn{2}{c}{RAVDESS}
& \multicolumn{2}{c}{VGGSound} \\
\cmidrule(lr){2-3}
\cmidrule(lr){4-5}
\cmidrule(lr){6-7}
\cmidrule(lr){8-9}
& A$\rightarrow$V & V$\rightarrow$A
& A$\rightarrow$V & V$\rightarrow$A
& A$\rightarrow$V & V$\rightarrow$A
& A$\rightarrow$V & V$\rightarrow$A \\
\midrule
UCMKD
& 31.8s & 44.13s
& 34.8s & 63.6s
& 103.3s & 83.8s
& 748.2s & 692.3s \\
Feature KD 
& 14.7s & 16.3s
& 22.5s & 31.9s
& 59.4s & 53.0s
& 624.8s & 237.9s \\
Relative overhead
& $2.16\times$ & $2.76\times$
& $1.54\times$ & $1.99\times$
& $1.73\times$ & $1.58\times$
& $1.20\times$ & $2.91\times$ \\
\bottomrule
\end{tabular}
}
\end{table}

\paragraph{Theoretical complexity.}
We now analyze the training complexity of UCMKD compared with feature-based KD. Let $N$ denote the number of training samples. We assume the same student and teacher backbones across methods. Let $F_s$ and $F_t$ denote the forward costs of the student and teacher models, respectively, and let $B_s$ denote the backward cost of the student model. Let $W(N)$ denote the cost of computing the Wasserstein distance, e.g., using Sinkhorn iterations with entropic regularization~\citep{peyré2020computationaloptimaltransport}. For feature-based KD, the per-epoch cost can be written as
\begin{equation}
C_{\mathrm{FKD}}=N(F_s + F_t + B_s) + W(N).
\end{equation}
For UCMKD, we perform one feature-alignment(FA) step and one label-alignment(LA) step per iteration, as used in our experiments. This requires additional student forward/backward passes, while the teacher only needs to be evaluated once. The resulting cost is
\begin{equation}
C_{\mathrm{UCMKD}}=3N(F_s + B_s) + NF_t + W(N).
\end{equation}
Therefore,
\begin{equation}
C_{\mathrm{UCMKD}}<3C_{\mathrm{FKD}},
\end{equation}
This shows that UCMKD introduces at most a constant-factor overhead compared with feature-based KD, rather than changing the asymptotic dependence on $N$. Importantly, the optimal transport term $W(N)$ is preserved rather than multiplied by the bilevel procedure. Thus, the main dataset-size-dependent alignment cost remains unchanged.  Consequently, UCMKD preserves the scalability of KD-style training while providing substantially improved performance.

\begin{table}[t]
\centering
\caption{Scalability evaluation using a ViT-based architecture with ViT-L (ViT-L/16, ~300M+ parameters) as the teacher and ViT-S (patch16-224, ~22M parameters) as the student.}
\label{tab:vit_large_results}
\begin{tabular}{lcccc}
\toprule
\multirow{2}{*}{Method} 
& \multicolumn{2}{c}{AVE} 
& \multicolumn{2}{c}{RAVDESS} \\
\cmidrule(lr){2-3} \cmidrule(lr){4-5}
& A$\rightarrow$V & V$\rightarrow$A 
& A$\rightarrow$V & V$\rightarrow$A \\
\midrule
Teacher         & 72.64 & 76.87 & 95.50 & 90.57 \\
CE              & 51.19 & 53.73 & 65.63 & 66.13 \\
Feature KD & 56.47 & 54.48 & 74.53 & 62.24 \\
\textbf{UCMKD}  & \textbf{59.45} & \textbf{61.19} & \textbf{88.42} & \textbf{77.42} \\
\bottomrule
\end{tabular}
\end{table}

\begin{table}[t]
\centering
\caption{Comparison with recent feature-based KD baselines in the unpaired cross-modal setting.}
\label{tab:modern_kd_baselines}
\begin{tabular}{lcccccc}
\toprule
\multirow{2}{*}{Method}
& \multicolumn{2}{c}{AVE}
& \multicolumn{2}{c}{RAVDESS}
& \multicolumn{2}{c}{CREMA-D} \\
\cmidrule(lr){2-3}
\cmidrule(lr){4-5}
\cmidrule(lr){6-7}
& A$\rightarrow$V & V$\rightarrow$A
& A$\rightarrow$V & V$\rightarrow$A
& A$\rightarrow$V & V$\rightarrow$A \\
\midrule
NORM   & 27.68 & 50.49 & 64.97 & 70.12 & 70.16 & 60.48 \\
REVIEW & 27.12 & 46.27 & 62.61 & 68.42 & 69.76 & 60.62 \\
\textbf{UCMKD}
& \textbf{34.16} & \textbf{52.24}
& \textbf{73.83} & \textbf{74.43}
& \textbf{71.64} & \textbf{66.67} \\
\bottomrule
\end{tabular}
\end{table}

\section{Implementation detail and Hyperparameters}
\label{sec:hyperparams_deatail}
In this section, we provide more implementation details and hyperparameters to reproduce our empirical results. To ensure fair comparisons, we adopt the same hyperparameters for all compared baselines. The specific hyperparameters are provided in Table~\ref{tab:hyperparams}. All experiments are run on an NVIDIA RTX A6000 GPU, and results are averaged over 5 independent runs. Our official implementation can be found at~\url{https://github.com/Duckduck-05/UCMKD}.

\begin{table}[t]
\centering
\caption{Hyperparameter configurations for multimodal datasets \textbf{AVE}, \textbf{CREMA-D}, \textbf{RAVDESS}, \textbf{VGGsound}. FA epoch and LA epoch denote the number of epochs minimizing Feature Alignment(\textbf{FA}) and Label Alignment(\textbf{LA}) in a single distillation epoch. $\lambda_1$ and $\lambda_2$ are the weights of Feature Alignment loss and Label Alignment loss in Algorithm~\ref{algo:metaKD}.}
\label{tab:hyperparams}
% \vskip 0.15in
\begin{footnotesize} % Slightly smaller than \small to ensure fit
\setlength{\tabcolsep}{5pt} % Adjusts spacing between columns
\begin{tabular}{lcccc}
\toprule
Hyperparameter & \textbf{AVE} & \textbf{CREMA-D} & \textbf{RAVDESS} & \textbf{VGGsound} \\
\midrule
Backbone      & ResNet-18     &ResNet-18       & ResNet-18    & ResNet-18   \\
Batch size    & 64       & 64        & 64      & 64     \\
Epoch         & 100       & 100        & 100      & 100    \\
Optimizer     & SGD      & SGD     & SGD    & SGD   \\
Learning rate & 1e-2     & 1e-2     & 1e-2    & 1e-2   \\
\midrule
$\lambda_1$      & 1      & 1       & 1     & 1    \\
$\lambda_2$      & 1      & 1       & 1     & 1    \\
FA epoch      & 1       & 1        & 1      & 1     \\
LA epoch     & 1        & 1         & 1       & 1      \\
\bottomrule
\end{tabular}
\end{footnotesize}
\end{table}
\section{Additional Experimental Results and Ablation Studies}
\label{sec:addtional_exp_results}
In this section, we present detailed experimental results, including standard deviations, and additional ablation studies of the proposed method that support the empirical analysis in Section~\ref{sec:empirical_analysis}.

\begin{table}[t]
\caption{Prediction accuracy with standard deviation on RAVDESS and CREMA-D with backbone ResNet-50 across different unpaired setting baselines: Cross Entropy, Feature KD, our method, and paired setting Vanilla KD~\citep{hinton2015distillingknowledgeneuralnetwork}. Our method achieves the best performance on 3 out of 4 tasks.}
\label{tab:res50_comparison}
\centering
\begin{tabular}{lcccc}
\toprule
Method & \multicolumn{2}{c}{RAVDESS} & \multicolumn{2}{c}{CREMA-D} \\
\cmidrule(r){2-3} \cmidrule(l){4-5}
& $A \to V$ & $V \to A$ & $A \to V$ & $V \to A$ \\
\midrule
Teacher         & 63.64 & 70.13 & 65.46 & 74.06 \\
\midrule
CE   & 52.25 $\pm$ 0.8 & 57.58 $\pm$ 2.9 & 71.33 $\pm$ 1.4 & 61.16 $\pm$ 0.4 \\
Feature KD & 50.25 $\pm$ 0.9 & 72.03 $\pm$ 2.9 & 69.71 $\pm$ 0.2 & 62.37 $\pm$ 0.2 \\
\midrule
Vanilla KD      & 66.67 $\pm$ 2.86 &  72.60 $\pm$ 1.56 & \textbf{73.20 $\pm$ 1.65} & 62.54 $\pm$ 0.17 \\
\midrule
\textbf{Ours}   & \textbf{70.83 $\pm$ 4.2} & \textbf{74.13 $\pm$ 3.5} & 
72.36 $\pm$ 1.2 & \textbf{66.94 $\pm$ 0.9} \\
\bottomrule
\end{tabular}
\end{table}

\begin{table}[t]
\centering
\caption{Performance under increasingly challenging unpaired settings with marginal mismatch, domain shift, and label imbalance on RAVDESS.}
\label{tab:distribution_mismatch}
\begin{tabular}{lcccccc}
\toprule
\multirow{2}{*}{Method}
& \multicolumn{3}{c}{RAVDESS A$\rightarrow$V}
& \multicolumn{3}{c}{RAVDESS V$\rightarrow$A} \\
\cmidrule(lr){2-4}
\cmidrule(lr){5-7}
& Easy & Medium & Hard
& Easy & Medium & Hard \\
\midrule
Feature KD
& 65.37 & 56.34 & 54.55
& 69.80 & 42.86 & 39.06 \\
CE
& 65.47 & 61.04 & 51.45
& 70.66 & 63.24 & 44.96 \\
\textbf{UCMKD}
& \textbf{73.83} & \textbf{67.64} & \textbf{64.23}
& \textbf{74.43} & \textbf{67.93} & \textbf{51.56} \\
\bottomrule
\end{tabular}
\end{table}

\paragraph{Scalability with larger backbones.} Table~\ref{tab:res50_comparison} reports the prediction accuracy on RAVDESS and CREMA-D using ResNet-50 as the backbone, further demonstrating that our method is not tied to a specific architecture and remains effective under a stronger convolutional network. Across the four transfer directions, our method achieves the best performance in three cases, showing consistent robustness across datasets and modality-transfer settings. To further evaluate scalability in more realistic settings, we conduct additional experiments using a ViT-based architecture (ViT-L as the teacher and ViT-S as the student), which is representative of modern large-scale vision models. As shown in Table~\ref{tab:vit_large_results}, UCMKD consistently achieves the best student performance across all datasets and transfer directions. These results show that UCMKD remains effective when moving from ResNet-based backbones to substantially larger ViT-based architectures. Together with the complexity analysis (See Appendix~\ref{sec:complexity}), this provides further evidence that our approach is scalable in realistic settings.

\paragraph{Compare with modern feature-based KD methods.} We further evaluate UCMKD against two recent feature-based knowledge distillation methods, REVIEW~\citep{Chen2021Review} and NORM~\citep{liu2023norm}. Although these methods were originally designed for paired data knowledge distillation, they can be adapted to the unpaired cross-modal setting and serve as strong feature-based baselines, as discussed in Section~\ref{sec:intro}. As shown in Table~\ref{tab:modern_kd_baselines}, UCMKD consistently outperforms both REVIEW and NORM across all datasets and transfer directions. These results further demonstrate that simply applying modern feature-based KD objectives is insufficient for the unpaired cross-modal distillation problem.

\paragraph{Robustness under distributional mismatch.}
We further evaluate UCMKD under increasingly challenging unpaired settings with distributional mismatch between the teacher and student modalities. Specifically, we consider three levels of difficulty: \emph{Easy}, which follows the random permutation setting used in the main text; \emph{Medium}, which introduces marginal mismatch by sampling teacher and student modalities from disjoint sample pools; and \emph{Hard}, which further incorporates modality-specific noise and label imbalance between the teacher and student datasets. As shown in Table~\ref{tab:distribution_mismatch}, both CE and Feature KD suffer substantial performance degradation as the mismatch becomes stronger. In contrast, UCMKD shows a more gradual decline and consistently achieves the best performance across all settings and transfer directions. These results demonstrate that UCMKD is more robust to marginal mismatch, domain shift, and label imbalance in realistic unpaired cross-modal distillation scenarios.

\end{document}